\PassOptionsToPackage{hyphens}{url}
\PassOptionsToPackage{table}{xcolor}
\documentclass{article}

\usepackage{iclr2027_conference,times}

\usepackage{graphicx}
\graphicspath{{Figures/}{appendix/figures/}}
\usepackage{wrapfig}
\usepackage{xcolor}
\usepackage{amsmath,amssymb}
\usepackage{array}
\usepackage{tabularx}
\usepackage{booktabs}
\usepackage{makecell}
\usepackage{pifont}
\usepackage{algorithm}
\usepackage{algorithmic}
\usepackage{listings}
\usepackage{placeins}
\usepackage{capt-of}
\usepackage{xparse}
\usepackage{textcomp}
\usepackage{tcolorbox}
\usepackage{tikz}
\usetikzlibrary{arrows.meta,positioning}
\tcbuselibrary{skins,breakable}
\usepackage{hyperref}
\usepackage{url}
\usepackage{enumitem}

\newcommand{\cmark}{\ding{51}}
\newcommand{\xmark}{\ding{55}}
\newcolumntype{C}{>{\centering\arraybackslash}X}
\newcolumntype{P}[1]{>{\raggedright\arraybackslash}p{#1}}
\newcolumntype{L}[1]{>{\raggedright\arraybackslash}p{#1}}

\definecolor{PromptBack}{HTML}{F7F8FB}
\definecolor{PromptFrame}{HTML}{6F7F91}
\definecolor{PromptTitle}{HTML}{E8ECF3}
\newtcolorbox{promptbox}[1]{%
  enhanced,
  breakable,
  width=0.92\textwidth,
  colback=PromptBack,
  colframe=PromptFrame,
  colbacktitle=PromptTitle,
  coltitle=black,
  fonttitle=\bfseries\small,
  fontupper=\normalsize,
  boxrule=0.45pt,
  arc=1.2pt,
  left=6pt,
  right=6pt,
  top=5pt,
  bottom=5pt,
  toptitle=3pt,
  bottomtitle=3pt,
  title={#1}
}
\newcommand{\promptblock}[2]{%
  \begin{promptbox}{#1}
  #2
  \end{promptbox}
  \vspace{0.35em}
}

\title{FraudSociety: A Generative Agent-Based Society for the Financial-Fraud Economy}

\author{
\normalfont
Keyang Chen$^{1,\dagger}$,
Mingxuan Jiang$^{1,\dagger}$,
Zeping Li$^{1}$,
Yiwen Zhao$^{1}$,
Yifan Wang$^{1}$, \\
Qiuyang Zhang$^{1}$,
Shuaicheng Qi$^{2}$,
Yicun Liu$^{1}$,
Zhixin Li$^{1}$,
Sen Liu$^{1}$, \\
Yinan Jing$^{1}$,
Guangnan Ye$^{1,*}$,
Hongfeng Chai$^{1}$ \\
\\
$^{1}$Fudan University, Shanghai, China \\
$^{2}$University of Illinois Urbana-Champaign, Urbana, IL, USA \\
{\small $^\dagger$Equal contribution.
$^*$Corresponding author.}
}

\iclrfinalcopy % Uncomment only for the accepted camera-ready version.

\begin{document}
\pagestyle{empty}
\maketitle

\begin{abstract}
Financial fraud caused over \$15.9 billion in losses in the United States in 2025, according to the Federal Trade Commission~\citep{ftc2026databook}.
Understanding such large-scale harm requires jointly modeling informational and financial dynamics. Yet existing simulators typically capture only one side of this interaction, leaving a gap in modeling how fraud unfolds in the real world.
To bridge this gap, we introduce \textsc{FraudSociety}, the first LLM-agent-based simulator that jointly models these coupled dynamics. It integrates a dynamic social environment for information exposure and communication, persistent financial ledgers that preserve payment and loss histories, and a Need-Intent-Action architecture for civilian decision making. Fraudsters further adapt their communication and financial actions to individual interaction histories and ledger states, forming a closed feedback loop between the two dynamics.
We validate \textsc{FraudSociety} across macro-level outcomes, micro-level behavior, and intervention effects. At the macro level, the simulated fraud outcomes closely match real-world patterns across multiple key statistics. At the micro level, it reproduces strong loss concentration, with the top 10\% of victimization records accounting for 81.1\% of total losses, consistent with Canadian Anti-Fraud Centre statistics~\citep{cafc2025annual}. Counter-fraud experiments further reveal variation across intervention strategies, with the best-performing personalized intervention reducing the repeat-payment rate from 49.4\% to 10.9\%.
These results establish \textsc{FraudSociety} as both a high-fidelity simulator and a scalable testbed for evaluating anti-fraud strategies under diverse intervention conditions, thereby advancing research on financial fraud and its prevention.
\end{abstract}

\NewDocumentCommand{\zeping}
{ mO{} }{\textcolor{blue}{\textsuperscript{\textit{Zeping}}\textsf{\textbf{\small[#1]}}}}

\section{Introduction}
\label{sec:intro}
Financial fraud has become a large-scale socio-economic harm: in 2025, the U.S. Federal Trade Commission (FTC) received 3 million fraud reports with losses exceeding \$15.9 billion~\citep{ftc2026databook}.
In practice, financial fraud unfolds through tightly coupled \textit{informational} and \textit{financial} dynamics, with each continuously affecting the other throughout the process.
Fraudulent information spreads through public content, platform recommendations, and private communication, gradually shaping a victim's attention, trust, and willingness to pay~\citep{anderson2024exploring,wang2026scoping}. As fraudulent information spreads, victims may make an initial payment, transfer additional funds, or attempt to withdraw purported returns.
These financial events, in turn, reshape subsequent communication: Fraudsters may provide small returns to build trust, solicit further transfers after an initial payment, or introduce new excuses to delay withdrawals~\citep{anderson2024exploring}. Financial loss therefore emerges from an interaction in which informational dynamics shape financial actions and evolving financial states reshape subsequent communication~\citep{oak2025hello}.

Studying financial fraud therefore requires modeling how informational and financial dynamics interact at the individual level and collectively give rise to population-level outcomes. Agent-based modeling (ABM) provides a natural computational paradigm for capturing this micro-to-macro process~\citep{epstein1999agent}.
Existing approaches, however, typically capture only one side of this coupling.
Traditional fraud-oriented ABMs focus on \textit{financial dynamics}, simulating payment networks and fraudulent transactions without modeling the informational processes~\citep{oztas2023enhancing}.
In contrast, LLM-based ABMs focus on \textit{informational dynamics}, supporting adaptive social behavior but rarely linking these interactions to persistent financial states and transaction processes~\citep{yang2024oasis,piao2025agentsociety}. Consequently, both paradigms fail to capture this interaction, limiting their ability to model how real-world fraud unfolds over time.

% \newpage
% \begin{wrapfigure}{r}{0.50\textwidth}
% \centering
% \includegraphics[width=\linewidth]{Figures/moti.png}
% \caption{Existing simulators fall short of real-world fraud by capturing only informational or financial dynamics. \textsc{FraudSociety} closes this gap by jointly modeling both.}
% \label{fig1}
% \end{wrapfigure}

\begin{figure*}[t]
\centering
\includegraphics[width=\textwidth]{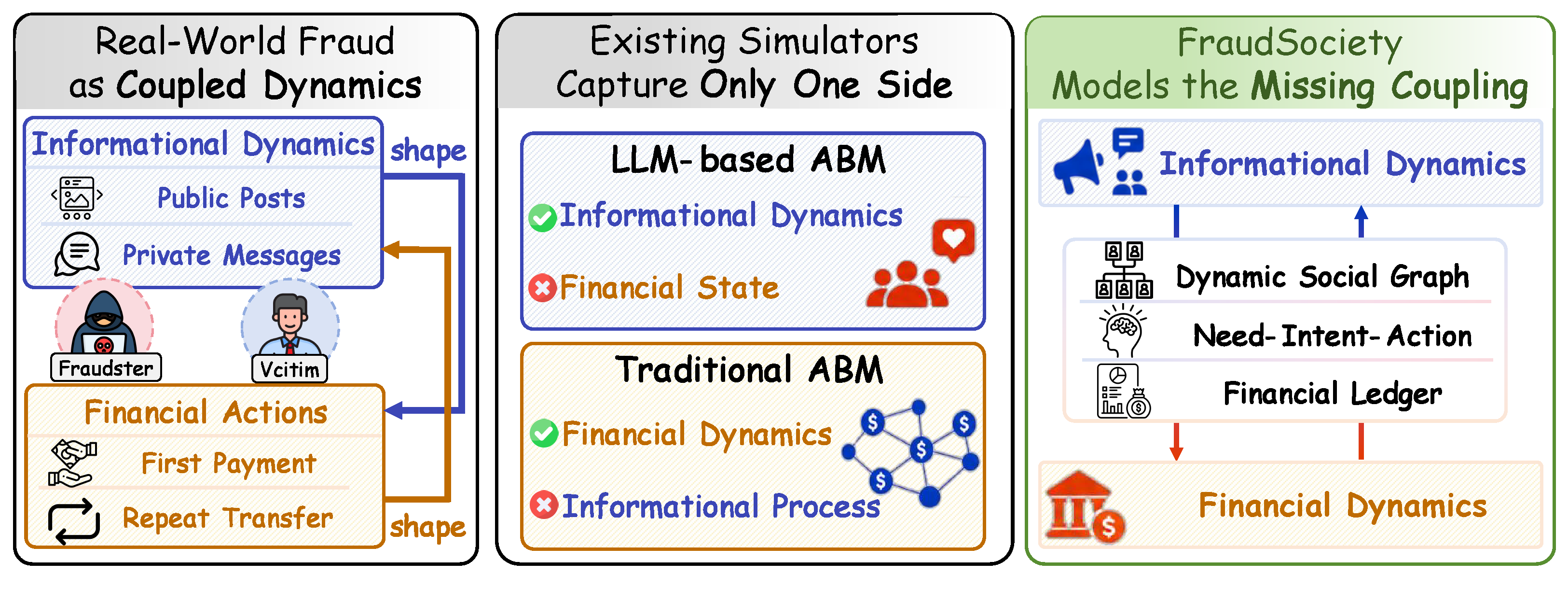}
\caption{Existing simulators fall short of real-world fraud by capturing only informational or financial dynamics. \textsc{FraudSociety} closes this gap by jointly modeling both.}
\label{fig1}
\end{figure*}

We introduce \textsc{FraudSociety}, an LLM-agent-based financial-fraud simulator that jointly models \textit{informational} and \textit{financial} dynamics. To model informational dynamics, we construct a dynamic social graph and adopt a Reddit-style ranking mechanism to govern public exposure, information diffusion, and private follow-up. To capture financial dynamics, we maintain persistent ledgers that track payments, returns, and withdrawal requests throughout each victimization process.
We then couple these dynamics through a Need-Intent-Action architecture, in which each individual's evolving needs and observed context shape behavioral intents and subsequent social and financial actions. These actions and their outcomes are recorded in persistent ledgers and become part of subsequent observations, while fraudsters adapt their communication and financial actions to each victim's response history and evolving financial state. Together, these processes form a closed feedback loop between informational and financial dynamics as interactions unfold over time.

By more closely approximating the structure of real-world fraud, \textsc{FraudSociety} serves as a high-fidelity simulator of the fraud process and provides a controlled testbed for counter-fraud interventions, enabling the systematic evaluation of the effects of different interventions on payment behavior and financial loss. To evaluate the fidelity and practical utility of \textsc{FraudSociety}, we focus on three specific research questions spanning multiple levels:
\textbf{RQ1:} At the macro level, can \textsc{FraudSociety} generate aggregate loss distributions consistent with stylized facts?
\textbf{RQ2:} At the micro level, is \textsc{FraudSociety} able to generate realistic victimization processes?
\textbf{RQ3:} From a practical perspective, can \textsc{FraudSociety} support the evaluation of counter-fraud interventions?

Through our experiments, we obtain consistent results across the macro, micro, and intervention levels.
At the macro level, \textsc{FraudSociety} achieves a Quartet Distance of 0.2978, indicating close alignment with multiple regulator-grounded stylized facts.
At the micro level, \textsc{FraudSociety} reproduces realistic behavioral heterogeneity and loss concentration, with the top 10\% of victimization records accounting for 81.1\% of total loss, broadly consistent with the statistics reported by the Canadian Anti-Fraud Centre (CAFC).
Finally, our counter-fraud experiments compare five intervention strategies and reveal substantial variation in their effectiveness. The best-performing personalized intervention reduces the repeat-payment rate from 49.4\% to 10.9\%, demonstrating the simulator's practical value for exploring and comparing real-world fraud interventions.

Collectively, our contributions are threefold:

\begin{itemize}[leftmargin=*]
\item We introduce \textsc{FraudSociety}, the first LLM-agent-based simulator to jointly model the tightly coupled informational and financial dynamics.

\item We propose a cross-level validation framework to systematically analyze both macro-level fidelity and underlying micro-level victimization mechanisms.

\item We show that \textsc{FraudSociety} can serve as a counter-fraud testbed, providing a scalable foundation for the design and analysis of real-world fraud prevention strategies.
\end{itemize}

\section{Related Work}
\label{sec:related}

\paragraph{Agent-based social simulation.}
Agent-based modeling (ABM) studies how heterogeneous individual behaviors and local interactions give rise to aggregate social and economic outcomes~\citep{epstein1999agent}. Traditional ABMs typically represent agents through predefined behavioral rules, enabling controlled analysis of emergence but limiting the complexity and adaptability of individual decision-making. Recent work incorporates LLM-based agents with profiles, memory, natural-language interaction, and adaptive behavior. Generative Agents models persistent individual routines and social interactions, while OASIS~\citep{yang2024oasis}, AgentSociety~\citep{piao2025agentsociety}, and SocioVerse~\citep{zhang2025socioverse} extend LLM-agent simulation to large-scale social environments. Recent studies further examine behavioral heterogeneity and adaptation, such as whether heterogeneous behavioral drivers lead to theoretically consistent strategy changes in simulated financial markets~\citep{li2026behavioral}. These studies establish the foundation for LLM-based simulation across settings, but primarily target general social or market dynamics rather than financial-fraud victimization.

\paragraph{Traditional simulation for financial fraud.}
Traditional financial-fraud simulators largely focus on transactions and capital flows. BankSim~\citep{lopez2014banksim} generates synthetic bank-payment behavior for fraud-detection research, while AMLSim~\citep{weber2018scalable} and SAML-D~\citep{oztas2023enhancing} model account interactions and laundering patterns for anti-money-laundering analysis. More recent systems such as TransXion~\citep{chen2026transxion} incorporate heterogeneous entities, demographic profiles, and structured transaction behaviors to construct realistic financial-crime graphs. These simulators provide persistent transaction records and controllable fraud labels, making them useful for detection and graph-learning benchmarks at larger scales. However, in these simulators, fraud-related transactions are generally generated from predefined transaction rules, without explicitly modeling the preceding information exposure, social interaction, and adaptive persuasion that influence individual payment decisions.

\paragraph{LLM-agent simulation for financial fraud.}
Recent studies use LLM agents to model fraud-related communication and interaction. ScamAgents~\citep{badhe2025scamagents} examines autonomous, multi-turn scam calls with context tracking and adaptive deception strategies. Most closely related, MultiAgent4Fraud~\citep{ren2025ai} simulates collaborative financial fraud across public posts and private interactions, and studies fraud success and agent collaboration. Nevertheless, existing systems primarily assess dialogue plausibility, fraud success, or harmful-agent capability, rather than coupling these interactions with persistent financial states through which payments, returns, withdrawals, and cumulative losses feed back into subsequent behavior. They also provide limited analysis of whether heterogeneous individual victimization pathways collectively reproduce population-level loss patterns observed in real-world fraud statistics. \textsc{FraudSociety} complements these studies by jointly modeling informational and financial dynamics and evaluating the resulting simulation at both the individual and aggregate levels.
\begin{figure*}[t]
\centering
\includegraphics[width=\textwidth]{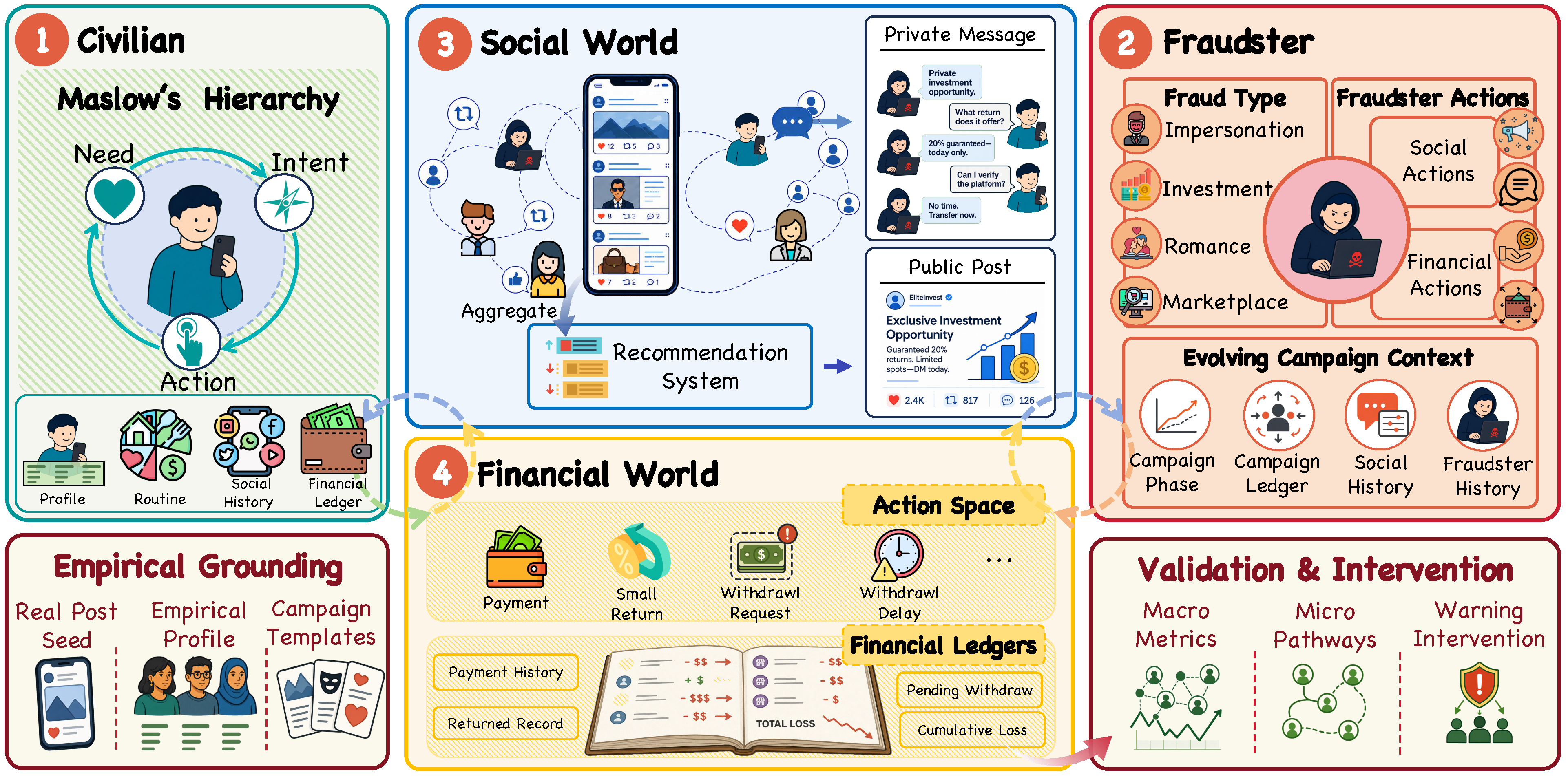}
\caption{Overview of \textsc{FraudSociety}. Civilians, guided by the Need-Intent-Action framework, act in Social and Financial Worlds, interacting with fraudsters who manage evolving campaign contexts.
Both sides take social and financial actions that update the environment and shape subsequent decisions. The simulation is initialized with real-world posts, empirical profiles, and campaign templates, and is validated through macro metrics, micro pathways, and warning interventions.}
\label{fig:framework}
\end{figure*}

\section{FraudSociety}\label{sec:fraudsociety}
\subsection{Overall Design of FraudSociety}\label{sec:method}

\textsc{FraudSociety} simulates fraud as a process coupling social interaction with financial consequences. As shown in Figure~\ref{fig:framework}, its simulation core consists of four components: Civilian, Fraudster, Social World, and Financial World. Civilians follow a \textbf{N}eed-\textbf{I}ntent-\textbf{A}ction (NIA) cycle based on their profiles, routine states, social histories, and financial records, while fraudsters manage persistent campaigns through adaptive social and financial actions. The Social World governs public information diffusion, personalized recommendation, and private communication, whereas the Financial World executes monetary actions and maintains persistent ledgers over repeated interactions.

At each hourly tick, the environment first updates routine, economic, platform, and campaign states and constructs agent-specific observations. Civilians and fraudsters then make decisions, after which the environment validates and executes their actions and commits the resulting social and financial state changes for subsequent decision cycles. 
We further ground the simulation with real-world post seeds, empirical civilian profiles, and fraud-campaign templates, and evaluate the resulting dynamics under diverse scenarios through macro-level metrics, micro-level pathways, and counter-fraud interventions. Details of the simulation cycle and state construction are provided in the Appendix.

\subsection{Civilian Modeling with Need-Intent-Action}\label{sec:method:nia}
In civilian modeling, the NIA architecture plays a central role in simulating how individuals respond to evolving social and financial environments across repeated interactions. As illustrated in Figure~\ref{fig:framework}, at each decision step, a civilian updates its needs based on the observed context, forms a behavioral intent, and selects a concrete action. The resulting feedback informs subsequent NIA cycles. The Appendix provides the formal definition and hourly agent cycle.

\textbf{Need.}
Need represents the dynamic internal state that shapes what a civilian prioritizes.
Inspired by Maslow's hierarchy of needs~\citep{maslow1943theory}, civilian needs are modeled along physiological, safety, and social dimensions.
As civilians engage in routine activities and receive social and economic signals, their need states evolve with their profiles and interaction histories, capturing heterogeneous priorities across civilians.
These evolving needs provide the internal basis for subsequent intent formation and, in turn, shape downstream behavior.
To track need dynamics and examine their micro-to-macro implications, \textsc{FraudSociety} records each civilian's need state at every step.

\textbf{Intent.}
Intent represents the behavioral objective that links an agent's current needs to executable behavior~\citep{ajzen1991theory,gollwitzer1999implementation}.
In \textsc{FraudSociety}, a civilian forms an intent by combining its current need state with the observed situation, selecting an activity route such as consumption, work, rest, or social interaction. The selected intent then restricts subsequent decision-making to a contextually relevant set of candidate actions, providing a structured transition from internal state to concrete behavior.
As the civilian receives new information and environmental feedback, its needs may change, leading the agent to revise its intent and adapt its subsequent actions accordingly.

\textbf{Action.}
Action represents the concrete behavior through which a civilian carries out its selected intent.
Conditioned on its current intent and observed context, the civilian selects actions from the corresponding action space.
For example, a social-interaction intent may lead to ordinary social behavior, fraud-campaign communication, or fraud-related payments.
Once these actions are executed, their consequences become part of subsequent observations, so that future decisions are informed by prior interactions and the current state.

\subsection{Fraudster Modeling}
\label{sec:method:operator}
In \textsc{FraudSociety}, fraudsters manage fraud campaigns with a context $\mathbf{x}$ that tracks social and financial interactions. For campaign $c$, decisions follow a hierarchical policy:
\begin{equation}
\label{eq:fraud-policy}
\begin{aligned}
q_{c,d+1}&\sim\pi_D(\cdot\mid q_{c,d},\kappa_c,\mathbf{x}_{c,d}),\\
(a^S_{c,i,t},a^F_{c,i,t})&\sim
\pi_H(\cdot\mid q_{c,d},\kappa_c,\mathbf{x}_{c,t},
\mathbf{h}_{c,i,t}),
\end{aligned}
\end{equation}
where $q_{c,d}$ is the daily campaign phase, $\kappa_c$ is the fraud type, and $\mathbf{h}_{c,i,t}$ is the interaction history with civilian $i$. $\pi_D$ and $\pi_H$ are structured-prompt LLM policies over admissible phase transitions and victim-specific actions, respectively.

\begin{wrapfigure}{r}{0.56\textwidth}
\centering
\includegraphics[width=\linewidth]{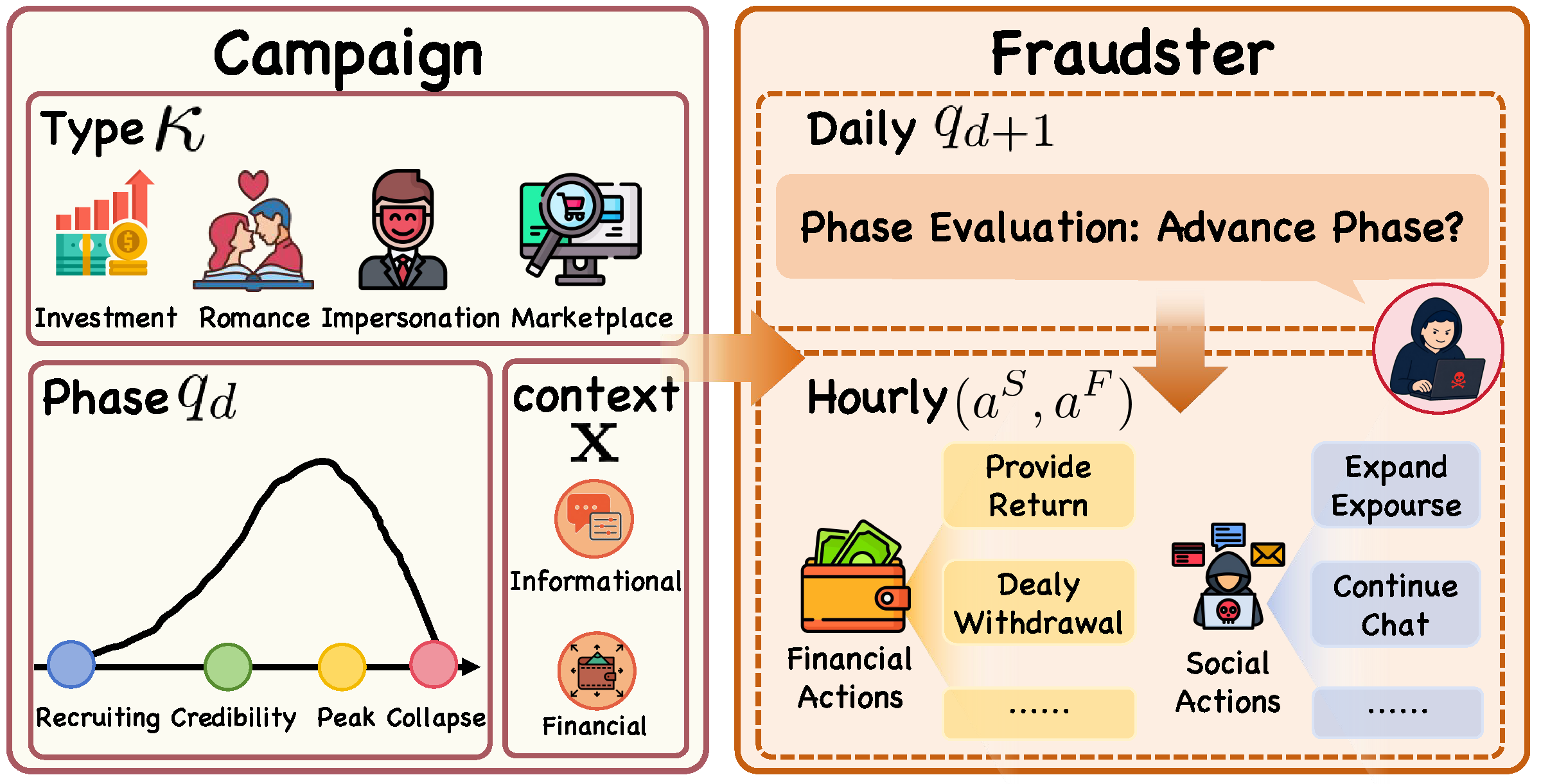}
\caption{Fraudster modeling in \textsc{FraudSociety}. Campaigns include a fraud type, evolving phase, and social-financial context, which together guide daily phase evaluation and victim-specific social and financial actions at the hourly level.}
\label{fig:fraud}
\end{wrapfigure}

\textbf{Campaign Type and Phase.}
As summarized in Figure~\ref{fig:fraud}, \textsc{FraudSociety} models investment, impersonation, romance, and marketplace fraud based on major reporting taxonomies~\citep{ftc2025databook,fbi2025ic3}, with templates tailored to each fraud type. Drawing on scam-lifecycle and victim-grooming studies~\citep{lacey2020application}, each campaign progresses through recruiting, credibility, peak, and collapse, corresponding to exposure expansion, legitimacy building, intensified extraction, and termination. At the daily timescale, $\pi_D$ selects whether to remain in the current phase or advance to the next, excluding backward and skipped transitions. The resulting phase and fraud type condition subsequent hourly decisions.

\textbf{Fraudster Action.}
At each hour, the fraudster selects victim-specific social and financial actions from the current campaign context and interaction history. The social action $a^S$ expands exposure, initiates contact, or continues a conversation, whereas the financial action $a^F$ provides a small return, handles a withdrawal request, or requests further payment. Equation~\eqref{eq:fraud-policy} therefore links daily campaign planning with victim-specific hourly adaptation.

\subsection{Social and Economic Environment}
\label{sec:method:environment}

To connect micro-level decisions with macro-level fraud outcomes, \textsc{FraudSociety} couples a dynamic social platform with an economic environment and persistent financial ledgers. The social platform governs public exposure, social amplification, and private communication, while the economic environment records the consequences of civilian and fraudster actions. This design captures how exposure timing, civilian circumstances, and fraudster responses jointly produce persistent financial feedback.

\textbf{Social Platform and Exposure.}
The social environment is represented as a typed dynamic graph
$G_t=(V,E_t,\rho)$,
where $V$ contains civilians and fraudsters, $E_t$ contains social edges at tick $t$, and $\rho : E_t \rightarrow \mathbb{R}$, where $\mathcal{R}$ is the set of relation types, such as family, colleague, neighbor, and online contact. Edges are initialized from civilian profiles, geographic proximity, and existing relations, and evolve through following, posting, reposting, commenting, liking, and direct messaging. Fraud exposure occurs through two channels: public diffusion through feeds and social interactions, and private follow-up through direct messages.

The platform constructs each civilian's feed using two-stage ranking. First, public posts are ranked by a Reddit-style hot score.
For post $p$ created at $t_p$, let $l_p$ and $d_p$ denote its numbers of likes and dislikes, respectively. We define the net vote as $v_p=l_p-d_p$. The corresponding hot score is
\begin{equation}
\label{eq:hot-score}
h(p)=\operatorname{sign}(v_p)\log_{10}(\max(|v_p|,1))
+\frac{t_p-t_0}{\tau_r},
\end{equation}
where $t_0$ is a reference timestamp and $\tau_r$ is a time-decay factor (e.g., in seconds)..

Globally ranked posts are combined with content from followed users and previously interacted authors to form a candidate pool. The candidates are then reranked by the cosine similarity between each post embedding and an embedding of the civilian's recently interacted content, and the highest-ranked posts enter the visible feed. Public exposure therefore depends jointly on global popularity, social connections, and individual interaction history, whereas private exposure depends on fraudster follow-up and prior civilian responses.

\textbf{Economic Environment and Ledgers.} The economic environment updates each civilian's daily activities and financial state over time. At each hourly tick, civilians engage in routine activities such as work, rest, and consumption, while income and expenditure events update their available balance. These evolving routine and financial states become part of the civilian's observation and provide the context for the subsequent Need-Intent-Action cycle.

\textsc{FraudSociety} maintains separate financial ledgers for civilians and fraud campaigns. Each civilian ledger records the individual's evolving financial state, while each campaign ledger records campaign-related financial interactions, including payments, returns, withdrawal requests, delays, and follow-up payment requests. The civilian-visible ledger states become part of subsequent observations, allowing later decisions to depend on current financial conditions as well as prior payments, received returns, unrecovered losses, and withdrawal experiences. Formal observation construction and ledger details are provided in the Appendix.

\definecolor{academicgreen}{HTML}{E8F0EC} 

\section{Experiments}
\label{sec:experiments}

We evaluate \textsc{FraudSociety} through three research questions that together form a cross-level validation chain: \textbf{RQ1} examines whether the simulator can reproduce aggregate loss distributions consistent with regulator-grounded stylized facts; \textbf{RQ2} examines whether these aggregate outcomes emerge from empirically grounded multi-step victimization pathways and whether key behavioral and economic mechanisms exhibit plausible patterns; and \textbf{RQ3} examines whether the simulator can serve as a counter-fraud testbed by capturing how different warning strategies affect payment behavior and financial loss. Correspondingly, our experiments proceed from macro-level aggregate validation, to micro-level pathway and mechanism analysis, and finally to intervention evaluation within the same simulated environment.

We instantiate \textsc{FraudSociety} with 112 civilians, 8 fraudsters, and 8 active campaigns over 480 hourly steps, using Qwen3-14B for all agent decisions. All variants and intervention conditions share the same civilian population, campaign initialization, and random seed to ensure controlled comparisons. Further details are provided in the Appendix.

\subsection{Macro-Level Validation}
\label{sec:exp-QD}

\textbf{Evaluation Metrics.} To evaluate macro-level realism, we introduce the Quartet Distance (QD), a regulator-grounded audit score cross-referencing simulated fraud losses against four empirical FTC stylized facts: Heavy-Tailed Loss concentration (HTL), Age-stratified Loss Severity profiles (ALS), Fraud-family Severity Hierarchy (FSH), and Age-specific Loss-Report shares (ALR). The composite score is:
{\small
\[
\mathrm{QD} = \frac{1}{N} \left( d_{\mathrm{HTL}} + d_{\mathrm{ALS}} + d_{\mathrm{FSH}} + d_{\mathrm{ALR}} \right),
\]
}
\par\noindent
where unavailable component distances are omitted from the sum, and $N$ denotes the number of included components.

Each component distance captures a distinct aspect of the simulated loss surface. 
$d_{\mathrm{HTL}}$ measures whether the simulation produces a comparable high-loss tail by comparing the simulated share of high-loss victims with the FTC target after rescaling the empirical high-loss threshold to the simulation scale. 
$d_{\mathrm{ALS}}$ compares the median cumulative loss profile across seven adult age bins, auditing whether age-stratified loss severity follows the FTC pattern. 
$d_{\mathrm{FSH}}$ evaluates whether the relative severity hierarchy between fraud families is preserved, using the investment-to-impersonation median-loss ratio as the primary cross-family severity contrast. 
$d_{\mathrm{ALR}}$ compares the age distribution of positive-loss victims with the FTC age-specific loss-report shares. 
Together, these four component distances audit whether the simulated fraud economy reproduces the coarse shape of regulator-observed victimization. We provide the full component definitions in the Appendix.

\begin{table*}[t]
    \centering
    \footnotesize
    \setlength{\tabcolsep}{1pt}
    \renewcommand{\arraystretch}{1.15}
    \caption{Comparison of baselines. \checkmark{} denotes support, whereas \ding{55} indicates unsupported mechanisms.}
    \begin{tabularx}{\textwidth}{
        P{2.5cm}
        *{6}{C}
    }
        \toprule

        \rule{0pt}{10pt}
        \textbf{System}
        & \makecell[c]{\textbf{Demographic}\\\textbf{profile}}
        & \makecell[c]{\textbf{Social}\\\textbf{exposure}}
        & \makecell[c]{\textbf{Private}\\\textbf{persuasion}}
        & \makecell[c]{\textbf{Financial}\\\textbf{state}}
        & \makecell[c]{\textbf{Repeated}\\\textbf{payment}}
        & \makecell[c]{\textbf{Withdrawal}\\\textbf{feedback}}
        \\

        \midrule

        \rule{0pt}{10pt}
        TransXion
        & \cmark
        & \xmark
        & \xmark
        & \cmark
        & \cmark
        & \xmark
        \\

        \rule{0pt}{10pt}
        MultiAgent4Fraud
        & \cmark
        & \cmark
        & \cmark
        & \xmark
        & \xmark
        & \xmark
        \\

        \rowcolor{academicgreen}
        \rule{0pt}{10pt}
        \textsc{FraudSociety}
        & \cmark
        & \cmark
        & \cmark
        & \cmark
        & \cmark
        & \cmark
        \\

        \specialrule{\heavyrulewidth}{0pt}{0pt}
    \end{tabularx}

    \label{tab:system_comparison}
\end{table*}

\textbf{Baseline Comparison.} We compare \textsc{FraudSociety} with two closely related financial-fraud simulation baselines:

(1) \textbf{TransXion}~\citep{chen2026transxion}: a profile-aware financial-transaction simulator that models structured entity behaviors and realistic transaction graphs.

(2) \textbf{MultiAgent4Fraud}~\citep{ren2025ai}: a multi-agent simulation system for studying financial-fraud behavior.

Table~\ref{tab:system_comparison} compares the three systems. TransXion focuses on profile-conditioned financial dynamics without modeling the preceding informational process, whereas MultiAgent4Fraud models interactions leading to the first fraud transfer but treats that transfer as the terminal outcome. In contrast, \textsc{FraudSociety} jointly models informational and financial dynamics throughout the victimization process.

Table~\ref{tab:qd_comparison} reports the regulator-grounded fidelity results across four stylized dimensions. \textsc{FraudSociety} achieves the lowest QD ($\mathrm{QD}=0.2978$), and consistently shows smaller discrepancies in loss concentration, age-conditioned severity, fraud-family severity, and victim-share patterns. These results indicate that the simulated outcomes more closely reproduce the joint statistical characteristics observed in regulatory data. Compared with TransXion, \textsc{FraudSociety} further explicitly models distinct fraud types, enabling direct evaluation of loss patterns across different fraud families.

% Keep the two macro-level tables together before the micro-level table.
\begin{table*}[t]
    \caption{Regulator-grounded fidelity comparison using Quartet Distance (QD) and its four constituent evaluation metrics. Lower values correspond to better alignment with real-world stylized facts, with \textsc{FraudSociety} achieving the best overall fidelity.}
    \centering
    \small
    \setlength{\tabcolsep}{6pt}
    \renewcommand{\arraystretch}{1.15}

    \begin{tabularx}{\textwidth}{l*{5}{>{\centering\arraybackslash}X}}
        \toprule
        \textbf{System}
        & \textbf{HTL$\downarrow$}
        & \textbf{ALS$\downarrow$}
        & \textbf{FSH$\downarrow$}
        & \textbf{ALR$\downarrow$}
        & \textbf{QD$\downarrow$} \\
        \midrule

        TransXion
        & 0.3722
        & 0.9168
        & --
        & 0.1793
        & 0.4894 \\

        MultiAgent4Fraud
        & 0.6211
        & 0.6586
        & 0.6017
        & 0.5243
        & 0.6014 \\

        \rowcolor{academicgreen}
        \rule{0pt}{10pt}
        \textsc{FraudSociety}
        & \textbf{0.0893}
        & \textbf{0.6513}
        & \textbf{0.3012}
        & \textbf{0.1495} 
        & \textbf{0.2978} \\

        \specialrule{\heavyrulewidth}{0pt}{0pt}
    \end{tabularx}

    \label{tab:qd_comparison}
\end{table*}

\begin{table}[t]
\caption{
Micro-level mechanism results.
(a) Association between civilian need states and fraud transfers: transfer hours exhibit higher need coordinates, and higher-need states show more frequent transfers.
(b) Mechanism ablation under the QD audit, where lower scores indicate greater realism.
}
\centering

% ================= Left =================
\begin{minipage}[t]{0.40\textwidth}
\vspace{0pt}
\centering
{\footnotesize\textbf{(a)} Need-state association}\par
\smallskip

\footnotesize
\setlength{\tabcolsep}{2.5pt}
\renewcommand{\arraystretch}{1.05}

\begin{tabular*}{\linewidth}{@{\extracolsep{\fill}}lcc@{}}
\toprule
Need &
\makecell[c]{Need coordinate\\All $\rightarrow$ transfer} &
\makecell[c]{Transfers / 100 h\\Low $\rightarrow$ high} \\
\midrule
Safety & 0.598 $\rightarrow$ 0.710 & 0.94 $\rightarrow$ 1.94 \\
Social & 0.630 $\rightarrow$ 0.712 & 1.03 $\rightarrow$ 1.91 \\
\bottomrule
\end{tabular*}
\end{minipage}
\hfill
% ================= Right =================
\begin{minipage}[t]{0.58\textwidth}
\vspace{0pt}
\centering
{\footnotesize\textbf{(b)} Mechanism ablation}\par
\smallskip

\footnotesize
\setlength{\tabcolsep}{3pt}
\renewcommand{\arraystretch}{1.05}
\setlength{\aboverulesep}{0pt}
\setlength{\belowrulesep}{0pt}

\begin{tabular*}{\linewidth}{@{\extracolsep{\fill}}lccccc@{}}
\toprule
\rule{0pt}{9pt}
Variant
& HTL$\downarrow$
& ALS$\downarrow$
& FSH$\downarrow$
& ALR$\downarrow$
& QD$\downarrow$ \\
\midrule

\rowcolor{academicgreen}
\rule{0pt}{9pt}
\textsc{FraudSociety}
& 0.0893 & 0.6513 & 0.3012 & 0.1495 & 0.2978 \\

\rule{0pt}{9pt}
w/o NIA
& 1.0000 & 0.6538 & 0.7908 & 0.1883 & 0.6582 \\

\rule{0pt}{9pt}
w/o Ledger
& 0.7005 & 0.5726 & 0.3410 & 0.1898 & 0.4510 \\

\bottomrule
\end{tabular*}
\end{minipage}

\label{tab:micro-level-results}
\end{table}

\begin{figure}[t]
\centering
\begin{minipage}[t]{0.49\textwidth}
\centering
\includegraphics[width=\linewidth]{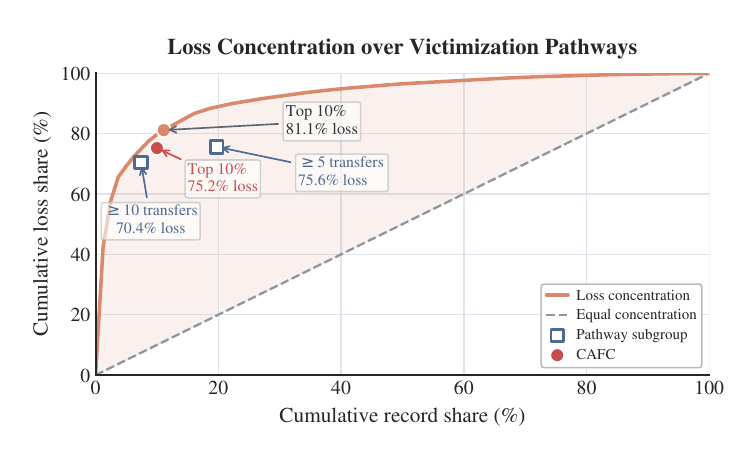}
\smallskip
{\small\textbf{(a)} Pathway loss concentration}
\end{minipage}
\hfill
\begin{minipage}[t]{0.49\textwidth}
\centering
\includegraphics[width=\linewidth]{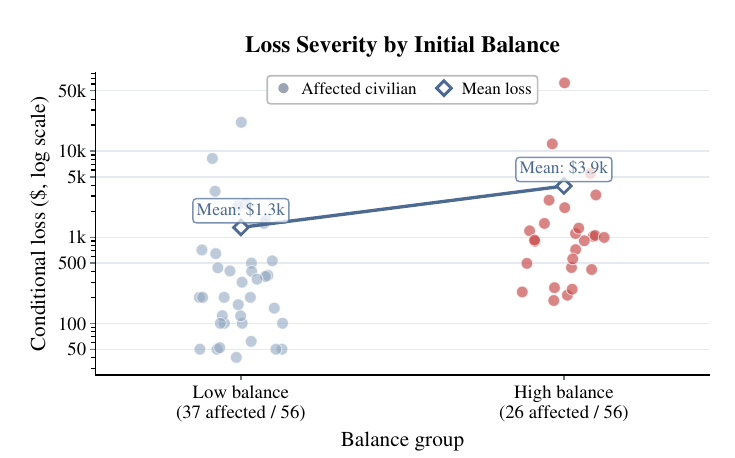}
\smallskip
{\small\textbf{(b)} Loss severity by initial balance}
\end{minipage}
\caption{Micro-level victimization outcomes. (a) Loss concentration across victimization pathways; pathway subgroups show concentration from repeated transfers, while the CAFC provides a real-world reference. (b) Loss severity across civilians with different initial balances; points show affected civilians, and markers show each balance group's mean loss.}
\label{fig:micro-level-outcomes}
\end{figure}

\subsection{Micro-Level Analysis}
\label{sec:exp-pathway-mechanism}
\textbf{Pathway Formation and Loss Escalation.}
\textsc{FraudSociety} exhibits a pathway-escalation pattern in which aggregate losses emerge through exposure, private contact, and repeated fraud transfers, consistent with cyber-victimization perspectives~\citep{cohen1979social,leukfeldt2016applying}. All 63 positive-loss victims experience campaign contact and private-DM exposure before their first transfer. The median positive-loss record contains 7 prior campaign contacts and a 72-hour first-contact-to-first-transfer lag, consistent with empirical evidence that fraud victimization develops through sustained exposure and contact~\citep{wang2026scoping}.

As shown in Figure~\ref{fig:micro-level-outcomes}(a), the cumulative loss curve lies above the equal-concentration baseline, indicating strong loss concentration. The top 10\% of records account for 81.1\% of total loss, broadly consistent with the 75.2\% top-10\% share reported by CAFC~\citep{cafc2025annual}. This concentration is closely associated with repeated transfers: records with at least five transfers account only 19.8\% of records but explain 75.6\% of total loss, while those with at least ten transfers still explain 70.4\%. These results show that \textsc{FraudSociety} reproduces both real-world loss concentration and its association with repeated transfers~\citep{xiao2025felt}.

\textbf{Individual Features in Fraud Victimization.}
We examine whether victimization outcomes vary with civilians' need states and economic conditions. As shown in Table~\ref{tab:micro-level-results}(a), fraud-transfer hours exhibit higher need-coordinate values than the overall distribution: safety need increases from 0.598 to 0.710, and social need increases from 0.630 to 0.712. Consistently, transfer rates rise from 0.94 to 1.94 between low- and high-safety-need states, and from 1.03 to 1.91 between low- and high-social-need states. These results show that \textsc{FraudSociety} exhibits patterns consistent with the state-dependent vulnerability in prior literature, where situational insecurity and unmet social needs can increase susceptibility to fraud-related decision making~\citep{chanenson2026didn}.

Economic conditions further shape victimizations. As shown in Figure~\ref{fig:micro-level-outcomes}(b), low-balance civilians are more likely to incur positive losses, reproducing the greater decision vulnerability associated with financial scarcity~\citep{shah2012some}. Among affected civilians, however, high-balance civilians incur larger conditional losses, consistent with routine-activity perspectives that higher-value targets will sustain greater losses once victimized~\citep{cohen1979social}. Together, these results show that \textsc{FraudSociety} reproduces theoretically grounded heterogeneity in both vulnerability to victimization and subsequent loss severity.

\textbf{Mechanism Ablation.}
We further test whether the proposed micro-mechanisms contribute to macro-level realism. Table~\ref{tab:micro-level-results}(b) compares the full model with two ablations: removing the \textit{Need-Intent-Action} architecture and disabling \textit{Financial Ledger}. The former removes need-conditioned civilian decision making, while the latter hides financial ledger states and feedback signals, such as income and cumulative investment.

Both ablations increase QD, indicating weaker alignment with regulator-grounded stylized facts. Since lower QD reflects better macro-level realism, these increases suggest that the \textit{Need-Intent-Action} architecture and \textit{ledger feedback} help bridge micro-level victimization pathways with aggregate loss patterns observed in real-world statistics.

\begin{figure}[t]
\centering
\includegraphics[width=\textwidth]{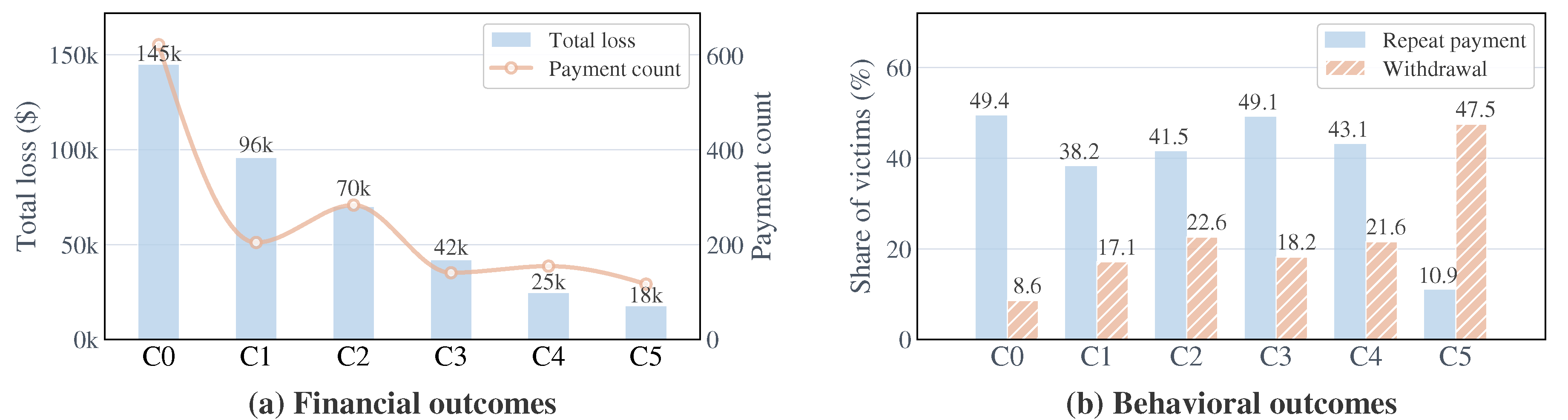}
\caption{Intervention effects on losses, payments, repeat payments, and withdrawal attempts.}
\label{fig:warning-outcomes}
\end{figure}

\subsection{Counter-Fraud Intervention Analysis}
\label{sec:exp-intervention}

We compare five warning strategies with the no-warning reference (\textbf{C0}): Platform broadcasts (\textbf{C1}), Monitor labels (\textbf{C2}), Random warning comments (\textbf{C3}), Peer warnings (\textbf{C4}), and Oracle personalized warnings (\textbf{C5}). Figure~\ref{fig:warning-outcomes} reports total loss and payments alongside repeat-payment and withdrawal-attempt rates. All five strategies reduce total loss, but through different downstream mechanisms.

\textbf{Platform-broadcast warnings.}
A fixed anti-fraud reminder is sent to all civilians daily. Relative to C0, payments fall from 623 to 205, loss from \$145K to about \$96K, and repeat payment from 49.4\% to 38.2\%, while withdrawal attempts rise from 8.6\% to 17.1\%. Broad reminders reduce activity, but limited targeting constrains their effect on continued payment and loss.

\textbf{Monitor labels.}
Warning labels mark suspicious public posts. Payments fall from 623 to 284 and total loss declines, but repeat payment remains high. Thus, labels mainly act at exposure; after civilians enter private follow-up and make an initial payment, the warning is absent and repeated payment is less affected.

\textbf{Random warning comments.}
Warnings inserted under posts regardless of suspiciousness reduce payments from 623 to 141 and total loss to \$42,014, but leave repeat payment nearly unchanged at 49.1\%. They create broad caution and suppress some initial payments, yet do not reliably interrupt repeats because they lack specific risk cues and victimization context.

\textbf{Peer warnings.}
Civilian prompts encourage warning-oriented actions such as commenting on suspicious posts. Positive-loss victims fall from 63 to 44, total loss to \$24,752, and maximum positive loss to \$3,800, indicating that observation-conditioned warnings compress high-loss pathways more effectively than random comments.

\textbf{Oracle personalized warnings.}
This idealized intervention triggers after a civilian's first fraud payment and uses an LLM to personalize the warning to the civilian and campaign. Payments fall from 623 to 117 and repeat payment from 49.4\% to 10.9\%, while withdrawal attempts rise to 47.5\%. Timely personalization therefore acts as a strong post-payment brake, shifting victims toward exit or recovery-seeking behavior.

Overall, \textsc{FraudSociety} provides a controlled testbed that reveals how counter-fraud interventions reshape victimization trajectories and downstream losses.

\section{Conclusion}

We introduce \textsc{FraudSociety}, the first LLM-agent-based simulator to jointly model the informational and financial dynamics of fraud. It couples adaptive social interaction and persistent ledgers through Need-Intent-Action decision making, so exposure shapes financial actions and financial outcomes reshape later communication. Experiments reproduce regulator-grounded macro-level loss patterns and micro-level victimization pathways, while ablations show that Need-Intent-Action and ledger feedback connect individual processes with aggregate patterns. The simulator also enables controlled comparisons of counter-fraud interventions and their downstream effects. Together, these capabilities make \textsc{FraudSociety} a scalable foundation for realistic financial-fraud simulation and systematic counter-fraud research.

\subsubsection*{Reproducibility Statement}
The appendix specifies the simulation clock, population construction, model and intervention settings, evaluation units, artifact schemas, complete runtime prompt templates, and the provenance of every reported intervention and Quartet Distance run.

\FloatBarrier
\bibliographystyle{iclr2027_conference}
\bibliography{iclr2027_references,iclr2027_conference}

\clearpage
\appendix

% Native ICLR appendix integrated into the same document.
\section{Guide to the Appendix}
\label{app:guide}

This appendix documents the implementation and evaluation details deferred from the main paper. Table~\ref{tab:app-roadmap} links each deferred topic to its detailed treatment below. Subsequent sections provide formal definitions, additional diagnostics, and English translations of the runtime prompts required to audit the reported mechanisms.

\begin{table*}[t]
\centering
\small
\setlength{\tabcolsep}{6pt}
\renewcommand{\arraystretch}{1.12}
\begin{tabular}{p{0.22\textwidth}p{0.36\textwidth}p{0.32\textwidth}}
\toprule
Main-paper reference & Appendix detail & Location \\
\midrule
Overall design & Global hourly ordering, closed-loop state commit, and operator/civilian decision timing & Sections~\ref{app:implementation} and~\ref{app:cycle} \\
Fraudster agents & Campaign lifecycle, public outreach, active-dialogue follow-up, financial feedback, and execution gates & Section~\ref{app:operator-hierarchy} \\
Civilian NIA model & Formal need update, active-need selection, intent routing, action generation, and the NIA ablation & Section~\ref{app:nia} \\
Social-economic environment & Bounded observation construction, social exposure, financial ledger transitions, and surfaced-state ablation & Sections~\ref{app:implementation} and~\ref{app:schemas} \\
Counter-fraud evaluation & Exact intervention triggers, visible channels, information assumptions, and matched outcome accounting & Sections~\ref{app:intervention-implementation} and~\ref{app:outcomes} \\
Micro-level validation & Victim--campaign accounting, contact-to-payment latency, escalation depth, and family-level pathways & Section~\ref{app:pathway-diagnostics} \\
Quartet Distance & Statistical units, all four component definitions, raw diagnostics, and baseline-alignment protocol & Sections~\ref{app:outcomes}, \ref{app:qd}, and~\ref{app:baseline-alignment} \\
Reproducibility & Runtime configuration, condition switches, and artifact-level provenance & Section~\ref{app:reproducibility} \\
Runtime policies & English translations of one shared-environment, eight civilian, and four operator prompt templates and their JSON contracts & Section~\ref{app:prompts} \\
\bottomrule
\end{tabular}
\caption{Correspondence between topics deferred from the main paper and their detailed treatment in this appendix.}
\label{tab:app-roadmap}
\end{table*}

\section{Implementation and Experimental Setting}
\label{app:implementation}

\paragraph{Population and temporal setup.}
Each simulation advances in hourly ticks, and every reported run comprises 480
steps over 20 simulated days.  We fix a common start time of
08:00 on March 11, 2025 for C0--C5 and all QD runs.  All conditions use
seed 42 and the same fixed profile realization, controlling
population resampling, stochastic initialization, and calendar/payday timing.
The full profile pool contains 112 civilian profiles and 8 criminal-operator
profiles.

\paragraph{LLM backbone.}
All LLM-controlled decisions in the reported experiments use Qwen3-14B as the backbone model. The same backbone is used for the no-warning baseline, counter-fraud interventions, and ablation settings; consequently, comparisons do not conflate the experimental mechanisms with changes in model family. Runtime inference uses temperature $0$ and task-specific output-token limits. Structured decisions are requested in JSON format. The parser removes reasoning tags, repairs recoverable syntax errors, validates action names and arguments against the active schema, and rejects actions that fail execution-time grounding constraints. The simulator seed fixes profile, graph, and intervention sampling.

\paragraph{Grounded civilian profiles and routine context.}
Each civilian profile specifies demographic attributes, age, education, occupation, home and work areas of interest, initial balance, income, salary day, consumption tendency, working hours, and Big-Five personality traits. These fields define the economic, mobility, and behavioral context of the hourly agent process. The environment uses an urban-scale spatial substrate. Home and work areas are taken from the fixed profile pool, and points of interest within a 2-km radius provide routine consumption alternatives. Salaries, fixed expenses, point-of-interest purchases, and fraud payments are recorded in the economy ledger; fraud payments are marked as illegal transactions.

\paragraph{Profile generation and U.S. age alignment.}
Civilian profiles are generated once and then held fixed across all reported runs. The profile generator preserves stable simulation identifiers, names, communities, and map-linked home and work areas while resampling demographic and behavioral variables. The adult age sampler follows the U.S. Census ACS 2022 S0101 age-by-sex table, normalized over adults aged 20 years and older~\citep{uscensus2022acsS0101}. The resulting target bins are 20--29, 30--39, 40--49, 50--59, 60--69, 70--79, and 80+. Table~\ref{tab:app-age-profile} reports both the target sampling weights and the fixed realization of 112 civilians used in the 120-agent experiments. Holding this finite-sample realization constant ensures that intervention comparisons reflect warning conditions rather than differences in population composition.

\begin{table*}[t]
\centering
\small
\setlength{\tabcolsep}{6pt}
\renewcommand{\arraystretch}{1.12}
\begin{tabular*}{0.82\textwidth}{@{\extracolsep{\fill}}lrrr}
\toprule
Age bin & ACS adult target & Realized civilians & Realized share \\
\midrule
20--29 & 17.2\% & 18 & 16.1\% \\
30--39 & 17.9\% & 16 & 14.3\% \\
40--49 & 16.5\% & 12 & 10.7\% \\
50--59 & 16.5\% & 32 & 28.6\% \\
60--69 & 15.8\% & 14 & 12.5\% \\
70--79 & 11.2\% & 10 & 8.9\% \\
80+ & 5.0\% & 10 & 8.9\% \\
\bottomrule
\end{tabular*}
\caption{Target and realized age composition of the 112 civilians used in the main experiments. The ACS column reports the adult-population sampling target; the remaining columns report the fixed profile realization used across all experimental conditions.}
\label{tab:app-age-profile}
\end{table*}

\paragraph{Big-Five behavioral grounding.}
Each civilian is assigned scores for openness, conscientiousness, extraversion, agreeableness, and neuroticism on a 1--10 scale. The generator samples these dimensions independently of age from clipped normal distributions. Openness, conscientiousness, extraversion, and agreeableness have mean $5.5$ and standard deviation $1.8$; neuroticism has mean $5.0$ and standard deviation $1.8$. These scores are operationalized as behavioral covariates that condition social responses, warning uptake, and fraud-payment decisions. Table~\ref{tab:app-bigfive-profile} summarizes the construction and behavioral interpretation of each dimension.

\begin{table*}[t]
\centering
\footnotesize
\setlength{\tabcolsep}{4pt}
\renewcommand{\arraystretch}{1.12}
\begin{tabular}{p{0.17\textwidth}p{0.16\textwidth}p{0.17\textwidth}p{0.41\textwidth}}
\toprule
Trait & Sampling target & Realized mean & Behavioral interpretation in civilian decisions \\
\midrule
Openness & $5.5\pm1.8$ & 5.46 & Higher values increase curiosity toward unfamiliar products, new platforms, and novel earning opportunities. \\
Conscientiousness & $5.5\pm1.8$ & 5.54 & Higher values increase verification, delayed commitment, and caution before major payments. \\
Extraversion & $5.5\pm1.8$ & 5.41 & Higher values increase social sharing, public engagement, and sensitivity to peer activity. \\
Agreeableness & $5.5\pm1.8$ & 5.17 & Higher values increase trust, compliance pressure, and difficulty refusing friend-like or authority-like requests. \\
Neuroticism & $5.0\pm1.8$ & 5.10 & Higher values increase sensitivity to urgency, loss avoidance, and fear-of-missing-out cues. \\
\bottomrule
\end{tabular}
\caption{Big-Five sampling distributions and behavioral interpretations for the 112 civilians. Scores are sampled on a 1--10 scale and clipped to the valid range; realized means are computed for the fixed experimental population.}
\label{tab:app-bigfive-profile}
\end{table*}

\paragraph{Financial and online-post grounding.}
Initial balances are sampled from age-specific transaction-account anchors derived from the Federal Reserve 2022 Survey of Consumer Finances~\citep{federalreserve2023scf}, with lognormal dispersion and an occupation multiplier. Occupation, salary, work schedule, consumption tendency, and retirement status then determine routine affordability and daily mobility. Public social-media context is initialized from two online-post seed pools, each containing 280 short posts. These seeds cover benign finance, cybersecurity, marketplace, work, charity, and relationship themes, as well as suspicious lures. During the simulation, criminal operators generate new public posts and private messages conditioned on fraud family, campaign phase, recent engagement, and visible user signals. The seed pools therefore provide topical grounding rather than fixed scripts.

\paragraph{Fraud campaign initialization.}
The simulator instantiates four fraud-family templates corresponding to the families examined in the main paper: investment, impersonation, marketplace, and romance. In the eight-operator setting, two operators are assigned to each family. Each operator is associated with one active campaign and a set of collection merchant accounts. Campaigns progress through four lifecycle phases: recruiting, credibility building, peak extraction, and collapse. In LLM-controlled mode, the operator selects phase transitions, while the campaign state records lifecycle status and the state changes associated with terminal collapse.

\paragraph{Social platform and social graph.}
The social layer uses an OASIS-style platform interface supporting public posts, comments, likes, follows, reposts, and direct messages. Each feed view contains at most 10 public posts and 20 unread private messages. The recommender first computes the Reddit-style hot score in Equation~(2) of the main paper and constructs a candidate set from globally high-ranked posts and posts authored by followed or positively engaged users. Candidates are then reranked using the hot score, follow and author-affinity indicators, and nonnegative embedding similarity to content with which the civilian recently interacted. If the embedding model is unavailable, reranking uses only the non-embedding features.

The initial graph is a hybrid small-world network rather than a uniformly random graph. Construction begins with the $k=6$ geographically nearest civilian neighbors, after which candidate edges are independently rewired with probability $p=0.15$. Within-community ties may be strengthened by occupation, salary, or education homophily, and additional online-contact edges are sampled by degree-preferential attachment. Each operator is connected to as many as 10 degree-weighted civilian targets, placing campaign exposure within the same graph as ordinary social interaction. Edges are labeled as family, colleague, neighbor, or online contact and are used by the social-proof, following, forwarding, and peer-warning mechanisms.

\paragraph{Counter-fraud conditions.}
Table~\ref{tab:app-intervention-conditions} summarizes the intended controlled
design, in which the six conditions vary in warning channel and trigger.
All six conditions share the same March 11 start, population realization,
simulator seed, and 480-hour horizon.

\begin{table}[tb]
\centering
\footnotesize
\setlength{\tabcolsep}{4pt}
\renewcommand{\arraystretch}{1.03}
\begin{tabular}{@{}L{0.17\columnwidth}L{0.77\columnwidth}@{}}
\toprule
Condition & Channel, trigger, and intended mechanism \\
\midrule
C0 & \textbf{No warning.} Baseline exposure and persuasion dynamics. \\ \addlinespace[1pt]
C1 & \textbf{Platform private warning.} Broadcast every 24 simulated hours; broad awareness with low personalization. \\ \addlinespace[1pt]
C2 & \textbf{Public monitor annotation.} Newly created posts are screened at the exposure surface. \\ \addlinespace[1pt]
C3 & \textbf{Random peer-style comment.} Five percent of civilians are sampled each step, producing sparse socially distributed warnings. \\ \addlinespace[1pt]
C4 & \textbf{Endogenous peer warning.} Civilian planning can interrupt persuasion through peer communication. \\ \addlinespace[1pt]
C5 & \textbf{Oracle personalized warning.} A transaction-specific private warning follows each executed fraud payment; the first-payment message is explicitly distinguished. \\
\bottomrule
\end{tabular}
\caption{Counter-fraud intervention conditions in the controlled design.
All conditions use the common March 11 clock.  C5 is interpreted as an oracle upper bound.}
\label{tab:app-intervention-conditions}
\end{table}

\paragraph{Run artifacts and comparative interpretation.}
Each run records social interactions, economic transactions, victim--campaign
states, QD statistics, and decision traces in artifacts linked by stable
identifiers.  All experimental conditions share the same population
realization, simulator seed, March 11 start time, and 480-hour horizon.  All
contrasts are single-run descriptive comparisons rather than estimates of
population-level effects across independent seeds.  Table~\ref{tab:app-ablation-switches}
identifies the two information-to-decision links removed by the ablations.

\begin{table}[t]
\centering
\footnotesize
\setlength{\tabcolsep}{4pt}
\renewcommand{\arraystretch}{1.03}
\begin{tabular}{@{}L{0.30\columnwidth}L{0.64\columnwidth}@{}}
\toprule
Variant & Mechanism state and effect \\
\midrule
Full model & \textbf{NIA enabled; ledger feedback enabled.} Need-conditioned routing uses visible cumulative financial fields. \\ \addlinespace[2pt]
NIA ablation & \textbf{NIA disabled; ledger feedback enabled.} Need updates remain logged, but the route is sampled uniformly from the feasible set. \\ \addlinespace[2pt]
Ledger ablation & \textbf{NIA enabled; ledger feedback disabled.} Transfers remain executable and included in evaluation, but cumulative financial fields are withheld from civilian prompts. \\
\bottomrule
\end{tabular}
\caption{Controlled mechanism ablations. Population, campaign, platform, economic, and model settings are held fixed.}
\label{tab:app-ablation-switches}
\end{table}
\subsection*{Counter-Fraud Intervention Realization}
\label{app:intervention-implementation}

Interventions are realized exclusively through information that can become visible in a civilian's bounded social observation. With the exception of C5, condition membership is never supplied to the civilian planner as a hidden label. A warning affects behavior only if it appears as an unread private message, an appended post label, a visible comment, or a civilian-generated social action and is subsequently selected into the bounded view. Table~\ref{tab:app-intervention-implementation} specifies the exact trigger and information assumption for each condition.

\begin{table}[tbp]
\centering
\scriptsize
\setlength{\tabcolsep}{3pt}
\renewcommand{\arraystretch}{1.06}
\begin{tabular}{@{}L{0.14\columnwidth}L{0.80\columnwidth}@{}}
\toprule
Condition & Code-level trigger, visible mutation, and information assumption \\
\midrule
C0 & No warning insertion or warning-specific prompt affordance. \\
C1 & At $t>0$ divisible by 24, pseudo-user 9999 inserts one fixed unread DM for every civilian; no classifier or victim-specific evidence is used. \\
C2 & After each new post, an independent LLM returns binary JSON from its first 400 characters.  A flagged post receives an in-place \texttt{[system warning]} suffix; parse failure leaves it unchanged. \\
C3 & At every $t>0$, the sampler seeded by $42+t$ selects six civilians.  Each adds a generic warning beneath a uniformly selected post among the 50 most recent, irrespective of suspiciousness. \\
C4 & The ordinary civilian social planner may comment under a visible post, publish a warning post, or privately warn a visible user using only its bounded observation. \\
C5 & After each executed fraud payment, pseudo-user 9999 inserts a transaction-keyed personalized DM.  It uses true family, amount, cumulative payment, count, and message cues; first-payment wording is distinguished. \\
\bottomrule
\end{tabular}
\caption{Code-level realization of the six counter-fraud conditions. C1--C4 operate on ordinary social surfaces. C5 deliberately consumes simulator ground truth and is used only as an intervention upper bound.}
\label{tab:app-intervention-implementation}
\end{table}

\paragraph{Mutation order and channel separation.}
C1 and C3 insert their social records after the six environment spaces have advanced and before civilian steps begin. C2 is instead a post-creation hook: the submitted post is first committed and then screened, so every later reader sees the same annotated content when the monitor flags it. C4 uses the ordinary civilian action executor and therefore produces no privileged social record type. C5 is called only after a fraud payment has passed affordability, campaign, and duplicate-transaction gates and has been committed to the victim ledger.

\paragraph{Failure and deduplication semantics.}
Malformed or unavailable C2 classifier output is fail-open: the post remains unchanged and no warning is inferred. C1 and C3 use the centralized social database layer to preserve the same private-message and comment schemas as ordinary actions. C5 uses a transaction identifier as its deduplication key, so retries cannot create repeated copies for the same payment; a later payment has a different identifier and can produce a new warning. These rules make the visible intervention record, rather than an unobserved condition flag, the auditable treatment event.

\paragraph{Attention is not forced.}
Insertion does not imply that the civilian reads or obeys a warning. Private warnings enter the unread-message queue, public labels compete for one of ten feed slots, and warning comments must be among the recent comments surfaced with a visible post. The subsequent NIA and social policies may respond, defer, ignore, forward, or act against a warning. Intervention outcomes therefore combine delivery, bounded exposure, and endogenous response, except for C5's explicitly oracle-informed targeting.

\section{Hourly Simulation Cycle and Closed-Loop Coupling}
\label{app:cycle}

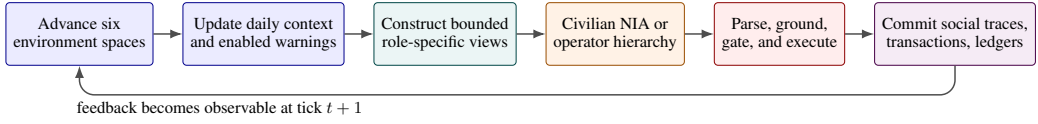
\begin{figure*}[tbp]
\centering
\resizebox{0.98\textwidth}{!}{%
\begin{tikzpicture}[
  node distance=5.5mm,
  >=Latex,
  every node/.style={font=\small,align=center},
  stage/.style={draw,rounded corners=2pt,minimum width=2.22cm,minimum height=1.18cm,inner sep=4pt},
  env/.style={stage,fill=blue!8,draw=blue!55!black},
  obs/.style={stage,fill=teal!8,draw=teal!55!black},
  agent/.style={stage,fill=orange!10,draw=orange!60!black},
  gate/.style={stage,fill=red!7,draw=red!55!black},
  state/.style={stage,fill=violet!8,draw=violet!55!black},
  flow/.style={->,thick,draw=black!70}
]
\node[env] (advance) {Advance six\\environment spaces};
\node[env,right=of advance] (triggers) {Update daily context\\and enabled warnings};
\node[obs,right=of triggers] (observe) {Construct bounded\\role-specific views};
\node[agent,right=of observe] (decide) {Civilian NIA or\\operator hierarchy};
\node[gate,right=of decide] (execute) {Parse, ground,\\gate, and execute};
\node[state,right=of execute] (commit) {Commit social traces,\\transactions, ledgers};
\draw[flow] (advance) -- (triggers);
\draw[flow] (triggers) -- (observe);
\draw[flow] (observe) -- (decide);
\draw[flow] (decide) -- (execute);
\draw[flow] (execute) -- (commit);
\draw[flow,rounded corners=8pt] (commit.south) -- ++(0,-0.55) -| node[pos=0.42,below,font=\footnotesize] {feedback becomes observable at tick $t+1$} (advance.south);
\end{tikzpicture}
}
\caption{Schematic of the within-hour execution order and closed feedback loop. Social exposure may induce financial actions during the current tick; committed messages, payments, returns, and withdrawal states enter subsequent agent observations.}
\label{fig:app-hourly-cycle}
\end{figure*}

Figure~\ref{fig:app-hourly-cycle} summarizes the update order described in the main paper. At the beginning of each hourly tick, the simulator advances the economic, information, campaign, mobility, event, and social-platform states. Daily macroeconomic context and condition-specific interventions are updated when their respective triggers are satisfied. Each agent then receives a bounded, role-specific projection of the resulting state and executes its planning and action process. Admissible social and financial effects are committed to the shared state before the next tick, ensuring that subsequent observations reflect the consequences of current-tick decisions.

Let $\mathcal{W}_t$ denote the shared social graph, public content, private-message store, economy, campaign ledgers, mobility state, and intervention state at tick $t$. For agent $i$, the observation builder $B_i$ produces a bounded view $o_{i,t}$ rather than exposing the full simulator state. Conditional on profile $p_i$, memory $m_{i,t}$, and role $r_i$, the agent policy produces a set of proposed actions,
\[
\tilde{\mathbf{a}}_{i,t}\sim
\pi_{r_i}\!\left(\cdot\mid p_i,m_{i,t},o_{i,t}\right).
\]
Parsing and execution constraints define a projection $\Gamma_{r_i}$ from proposed actions to the admissible action set. The shared-state transition is therefore
\[
\mathcal{W}_{t+1}
=
\mathcal{T}\!\left(
\mathcal{W}_{t},
\bigcup_{i\in\mathcal{A}}
\Gamma_{r_i}(\tilde{\mathbf{a}}_{i,t},o_{i,t})
\right),
\]
Agent memory is updated only from grounded execution results; actions with invalid targets, unaffordable payments, or unsupported operator operations therefore do not enter the persistent state.

\begin{algorithm}[t]
\small
\caption{Hourly simulation tick}
\label{alg:app-hourly-cycle}
\begin{algorithmic}[1]
\REQUIRE shared world state $\mathcal{W}_t$, agents $\mathcal{A}$, condition $c$
\STATE advance economy, information, campaign, mobility, event, and social spaces
\STATE update day-level context and fire warning triggers enabled by $c$
\FORALL{$i\in\mathcal{A}$ \textbf{in parallel}}
  \STATE construct bounded observation $o_{i,t}=B_i(\mathcal{W}_t)$
  \IF{$i$ is a civilian}
    \STATE synchronize income, routine, return, and withdrawal effects
    \STATE update needs; choose an available intent route; retrieve route-relevant memory
    \STATE plan and execute grounded routine, social, payment, or withdrawal actions
  \ELSE
    \STATE evaluate the campaign phase at its day-level decision point
    \STATE plan public outreach, active-dialogue follow-up, and financial feedback
    \STATE execute only actions that pass target, cadence, balance, and lifecycle gates
  \ENDIF
\ENDFOR
\STATE periodically synthesize operator reflections and export append-only traces
\RETURN committed shared state $\mathcal{W}_{t+1}$
\end{algorithmic}
\end{algorithm}

\begin{table*}[t]
\centering
\small
\setlength{\tabcolsep}{5pt}
\renewcommand{\arraystretch}{1.12}
\begin{tabular}{p{0.16\textwidth}p{0.29\textwidth}p{0.25\textwidth}p{0.20\textwidth}}
\toprule
Decision layer & Admissible action families & Principal execution gates & Persistent effects \\
\midrule
Civilian routine & work/rest, nearby consumption, no action & time slot, work schedule, nearby POI, available balance & mobility, legal transactions, routine memory \\
Civilian social & post, comment, like, repost, follow, private reply, ignore/defer, mark suspicious, forward warning & visible post and message identifiers, reply grounding, current route, social availability & graph edges, public traces, dialogue state, warning exposure \\
Civilian financial & fraud payment, hold, withdrawal request & visible campaign contact, positive affordable amount, campaign state, no duplicate pending request & economy transaction and victim-campaign ledger \\
Operator social & create post, feed-driven outreach, active-dialogue follow-up, like/repost/follow, coordination task & phase, visible eligible target, active-conversation membership, cadence, per-step deduplication & public exposure, private contact, task/shared memory \\
Operator financial & pay return, pay withdrawal, delay withdrawal, request top-up/unlock, no action & existing paid victim, pending request, available campaign pool, lifecycle consistency & return/payout transaction, delay state, follow-up message \\
\bottomrule
\end{tabular}
\caption{Action families and the grounding boundary between LLM proposals and persistent state changes.}
\label{tab:app-action-space}
\end{table*}

\section{Fraudster Hierarchical Decision Process}
\label{app:operator-hierarchy}

Fraudster behavior is implemented as a hierarchy of temporally distinct
policies rather than as a single hourly completion.  For campaign $c$, let
$z_{c,t}\in\{\textit{recruiting},\textit{credibility},\textit{peak},
\textit{collapse}\}$ denote its lifecycle phase.  The phase policy is evaluated
at most once per simulated date and its applied result is cached for the
remaining ticks of that date.  Execution enforces a monotone lifecycle:
proposed retreats are rejected, entry into \textit{peak} is rejected when the
campaign has neither a victim nor positive raised funds, and
\textit{collapse} is terminal.  Thus, if $\rho(\cdot)$ is the phase rank, the
committed transition satisfies
\[
\begin{aligned}
\rho(z_{c,t+1})&\geq\rho(z_{c,t}),\\
z_{c,t}=\textit{collapse}
&\Rightarrow z_{c,t+1}=\textit{collapse}.
\end{aligned}
\]
This projection separates strategic suggestions from lifecycle changes that
are structurally admissible.

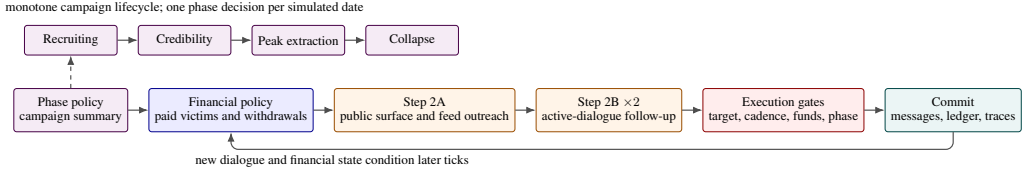
\begin{figure*}[t]
\centering
\resizebox{0.97\textwidth}{!}{%
\begin{tikzpicture}[
  >=Latex,
  every node/.style={font=\small,align=center},
  phase/.style={draw,rounded corners=2pt,minimum width=2.25cm,minimum height=0.72cm,fill=violet!8,draw=violet!55!black},
  policy/.style={draw,rounded corners=2pt,minimum width=2.38cm,minimum height=1.05cm,inner sep=4pt},
  pphase/.style={policy,fill=violet!8,draw=violet!55!black},
  pfinance/.style={policy,fill=blue!8,draw=blue!55!black},
  psocial/.style={policy,fill=orange!9,draw=orange!60!black},
  pgate/.style={policy,fill=red!7,draw=red!55!black},
  pstate/.style={policy,fill=teal!8,draw=teal!55!black},
  flow/.style={->,thick,draw=black!70}
]
\node[phase] (recruit) {Recruiting};
\node[phase,right=5mm of recruit] (cred) {Credibility};
\node[phase,right=5mm of cred] (peak) {Peak extraction};
\node[phase,right=5mm of peak] (collapse) {Collapse};
\draw[flow] (recruit) -- (cred);
\draw[flow] (cred) -- (peak);
\draw[flow] (peak) -- (collapse);
\node[font=\footnotesize,above=1.5mm of cred] {monotone campaign lifecycle; one phase decision per simulated date};

\node[pphase,below=8mm of recruit] (phasepol) {Phase policy\\campaign summary};
\node[pfinance,right=5mm of phasepol] (finance) {Financial policy\\paid victims and withdrawals};
\node[psocial,right=5mm of finance] (s2a) {Step 2A\\public surface and feed outreach};
\node[psocial,right=5mm of s2a] (s2b) {Step 2B $\times 2$\\active-dialogue follow-up};
\node[pgate,right=5mm of s2b] (gates) {Execution gates\\target, cadence, funds, phase};
\node[pstate,right=5mm of gates] (commitop) {Commit\\messages, ledger, traces};
\draw[flow] (phasepol) -- (finance);
\draw[flow] (finance) -- (s2a);
\draw[flow] (s2a) -- (s2b);
\draw[flow] (s2b) -- (gates);
\draw[flow] (gates) -- (commitop);
\draw[flow,dashed] (phasepol.north) -- (recruit.south);
\draw[flow,rounded corners=5pt] (commitop.south) -- ++(0,-0.42) -| node[pos=0.43,below,font=\footnotesize] {new dialogue and financial state condition later ticks} (finance.south);
\end{tikzpicture}
}
\caption{Fraudster decision hierarchy.  The daily lifecycle policy constrains
three lower-frequency or hourly policies.  Step~2A constructs public visibility
and feed-grounded outreach, whereas Step~2B only continues conversations in
which the civilian spoke last.  Proposed operations affect the shared state
only after deterministic execution gates.}
\label{fig:app-operator-hierarchy}
\end{figure*}

Figure~\ref{fig:app-operator-hierarchy} makes the separation of roles explicit:
the lifecycle layer restricts the strategies available on a given date, the
financial layer reacts to paid-victim and withdrawal states, Step~2A builds
public visibility and recruits new targets, and Step~2B is confined to active
dialogues.  The final deterministic gates project generative proposals onto
the executable action set.  The ordering therefore identifies both the policy
timescales and the point at which a proposal becomes persistent state.

The financial policy is active only in the \textit{credibility} and
\textit{peak} phases.  Its observation includes campaign funds, as many as
eight paid-victim summaries, pending withdrawal requests, latest received
messages, and recent payment or return events.  A change in this context
signature forces replanning.  Otherwise, a valid decision is reused for four
hours, shortened to two hours when a payment, return, or pending-withdrawal
event occurred within the previous two hours.  This event-sensitive cache
prevents financial feedback from becoming an artificial hourly heartbeat while
still allowing rapid responses to newly changed victim states.

The social policy is split into two non-interchangeable stages.  Step~2A may
create a post, engage with public content, follow users, and initiate
feed-grounded outreach.  Its prompt contains no more than ten visible
candidates, with at most six prompt slots reserved for untouched cold
profiles; the remaining slots can represent existing victims or warmed leads.
A stranger newly followed by the operator is not immediately eligible for a
direct message and must remain followed for six simulated hours unless already
registered as a campaign target.  Step~2A is also forbidden from messaging a
user whose conversation is active.  Step~2B is the complementary
active-dialogue policy: it receives no public-feed candidates, may address only
conversations in which the civilian spoke last, and is evaluated for at most
two follow-up rounds per hourly tick.  Public outreach and dialogue replies
have separate cadence clocks, so a reply does not consume the cold-outreach
window.

\begin{table*}[t]
\centering
\small
\setlength{\tabcolsep}{5pt}
\renewcommand{\arraystretch}{1.12}
\begin{tabular}{p{0.13\textwidth}p{0.20\textwidth}p{0.24\textwidth}p{0.33\textwidth}}
\toprule
Subpolicy & Invocation & Principal observation & Deterministic post-LLM constraints \\
\midrule
Lifecycle & At most once per simulated date & campaign age and phase, victims, raised and returned funds, outreach state, withdrawal pressure & valid phase label; monotone advance; nonempty economic evidence before peak; terminal collapse \\
Financial feedback & Credibility/peak; context change or 2--4 h cache expiry & campaign pool, paid-victim ledger, pending withdrawals, recent messages and financial events & existing paid victim or request; positive amount; available pool; valid lifecycle; no duplicate execution \\
Step 2A: public/outreach & When a phase-scaled outreach window is due or visible candidates exist & recent public posts, no more than 10 grounded candidates, warm and repeat leads, campaign and shared-task context & visible eligible target; no active dialogue; follow is Step~2A only; six-hour follow-to-DM warm-up for strangers; per-step deduplication \\
Step 2B: dialogue & Up to two rounds per tick when a civilian spoke last and reply cadence is due & bounded recent dialogue, account-state summary, campaign semantics, retrieved memory & target must belong to the active-conversation set; private-message action only; one sent message per target per tick \\
\bottomrule
\end{tabular}
\caption{Invocation scopes and execution constraints of the fraudster hierarchy.  The LLM selects among schema-valid proposals, while the final column is enforced by simulator code.}
\label{tab:app-operator-subpolicies}
\end{table*}

\section{Need-Intent-Action Formalization}
\label{app:nia}

For each civilian $i$ and hourly tick $t$, \textsc{FraudSociety} maintains the need vector
\[
\mathbf{s}_{i,t} =
\left(s^{\mathrm{satiety}}_{i,t},
s^{\mathrm{energy}}_{i,t},
s^{\mathrm{safety}}_{i,t},
s^{\mathrm{social}}_{i,t}\right) \in [0,1]^4,
\]
where larger values indicate greater need urgency, or equivalently lower need satisfaction. The default initial values are $0.62$, $0.15$, $0.10$, and $0.28$, respectively, and the activation thresholds are
\[
\boldsymbol{\theta}=(0.80,0.80,0.80,0.70).
\]
Let $E_{i,t}=\{k:s^k_{i,t}>\theta_k\}$. If $E_{i,t}$ is empty, the active-need variable is assigned the neutral sentinel $\bot$, corresponding to the implementation label \texttt{whatever}. Otherwise, the active need is selected from $E_{i,t}$ according to the priority order satiety, energy, safety, and social; ties within a priority level are resolved by greater urgency.

More explicitly, let $\mathbf{s}^{-}_{i,t}$ and $\mathbf{s}^{+}_{i,t}$ denote the pre-update and post-update need vectors. The structured need updater implements
\[
\mathbf{s}^{+}_{i,t}
=
\Pi_{[0,1]^4}
F_{\phi}\!\left(
\mathbf{s}^{-}_{i,t},
o_{i,t},
m_{i,t},
\ell_{i,t-1}
\right),
\]
where $o_{i,t}$ is the bounded observation, $m_{i,t}$ is retrieved recent memory, $\ell_{i,t-1}$ is the previous intent route, $F_{\phi}$ is the Qwen3-14B need-adjustment policy, and $\Pi$ denotes schema validation and range clipping. The update is sparse: coordinates omitted from a valid structured response remain unchanged. If no valid response is available, the complete vector is retained for that tick. In the notation above, $\mathbf{s}_{i,t}$ refers to this post-update vector. Active-need selection follows the lexicographic rule
\[
n_{i,t}=
\begin{cases}
\bot, & E_{i,t}=\varnothing,\\
\arg\min_{k\in E_{i,t}}\bigl(\operatorname{priority}(k),-s^k_{i,t}\bigr),
& \text{otherwise}.
\end{cases}
\]

Let $\mathcal{R}_{i,t}$ be the routes currently feasible under time, work, sleep, mobility, and social availability. The intent and action stages are
\[
\begin{aligned}
\ell_{i,t}
&\sim
\pi_I(\cdot\mid p_i,o_{i,t},\mathbf{s}_{i,t},n_{i,t},
\mathcal{R}_{i,t}),\\
\tilde{\mathbf{a}}_{i,t}
&\sim
\pi_A(\cdot\mid \ell_{i,t},o_{i,t},m_{i,t}).
\end{aligned}
\]
The selected intent route restricts the available action schema. For example, the social route exposes public and private communication actions, whereas the consumption route exposes nearby merchant choices. The execution projection $\Gamma$ in Section~\ref{app:cycle} removes ungrounded or infeasible actions before any state transition is committed.

The hourly civilian cycle proceeds as follows:
\begin{enumerate}
\item synchronize salaries, fixed expenses, campaign returns, and withdrawal payouts;
\item resolve sleep state and optionally perform an overnight check for unread messages;
\item update home-work mobility according to the civilian's schedule;
\item refresh the social feed, unread private messages, warning messages, recent comments, and active campaign contacts;
\item build a compact observation from recent memory, current balance, macro context, inbox state, visible posts, nearby points of interest, peer activity, and campaign ledger state;
\item update $\mathbf{s}_{i,t}$ through the need-adjustment LLM call, then derive the active need;
\item choose an intent route from the feasible set: social-media engagement, nearby consumption, or rest/work;
\item retrieve relevant narrative memories for the chosen route;
\item execute the route-specific action schema and update memory, social traces, economy transactions, and campaign ledgers.
\end{enumerate}

In the full model, the LLM selects the intent route conditional on the current need state and observation context. In the NIA ablation, need-update records are retained, but the civilian's primary route is sampled uniformly from $\mathcal{R}_{i,t}$. The ablation therefore removes need-conditioned route selection while preserving the social platform, campaign ledger, and economic routines.

\begin{figure*}[t]
\centering
\includegraphics[width=0.94\textwidth]{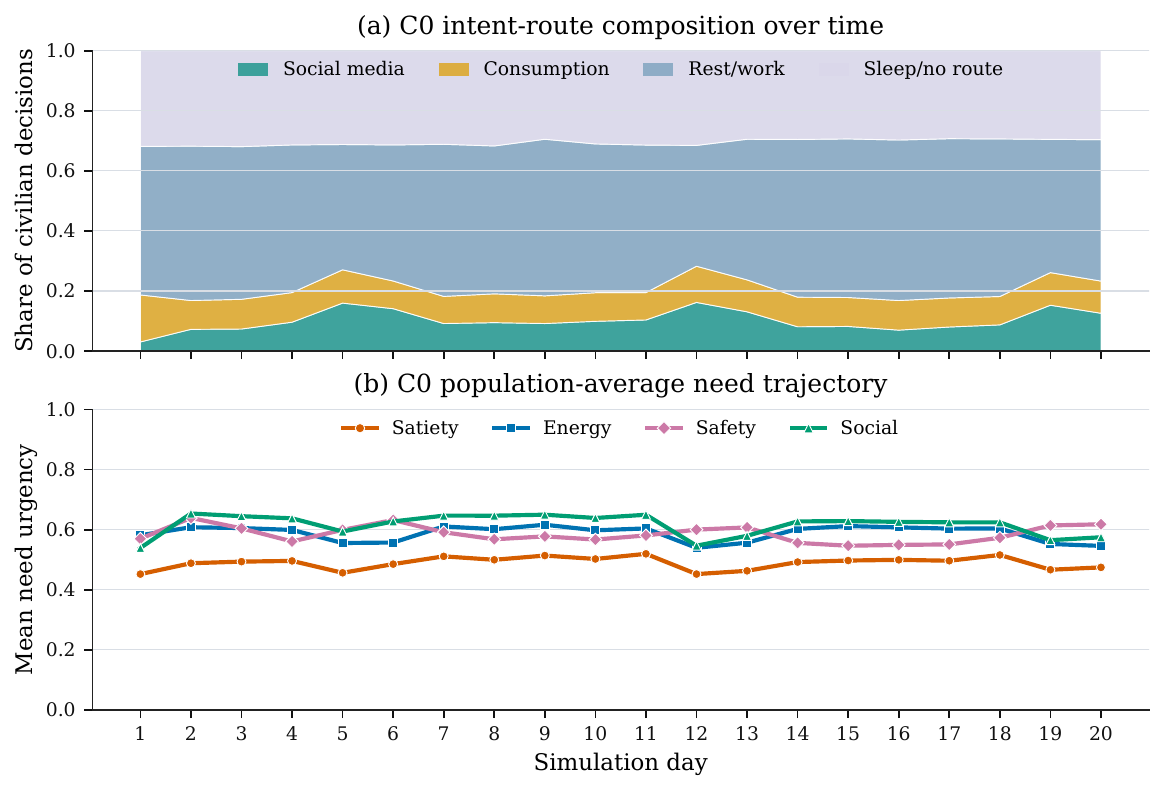}
\caption{C0 aggregate need-intent dynamics. Panel (a) reports the daily composition of civilian main activity routes under the no-warning baseline. Panel (b) reports the daily population-average values of the four need-urgency variables after need updates.}
\label{fig:app-c0-dynamics}
\end{figure*}

\paragraph{C0 aggregate need-intent dynamics.}
Figure~\ref{fig:app-c0-dynamics} provides a diagnostic summary of the no-warning baseline over the 20-day horizon. The route composition is dominated by routine behavior: rest/work and sleep/no-route account for most hourly decisions, whereas social-media and consumption routes exhibit smaller but persistent shares. Social need has the highest population-average urgency throughout most of the horizon, while satiety, energy, and safety remain more closely grouped and covary with the daily routine. This pattern is consistent with the intended role of need-conditioned intent in coupling routine activity with periodic social engagement.

\begin{figure*}[t]
\centering
\includegraphics[width=0.94\textwidth]{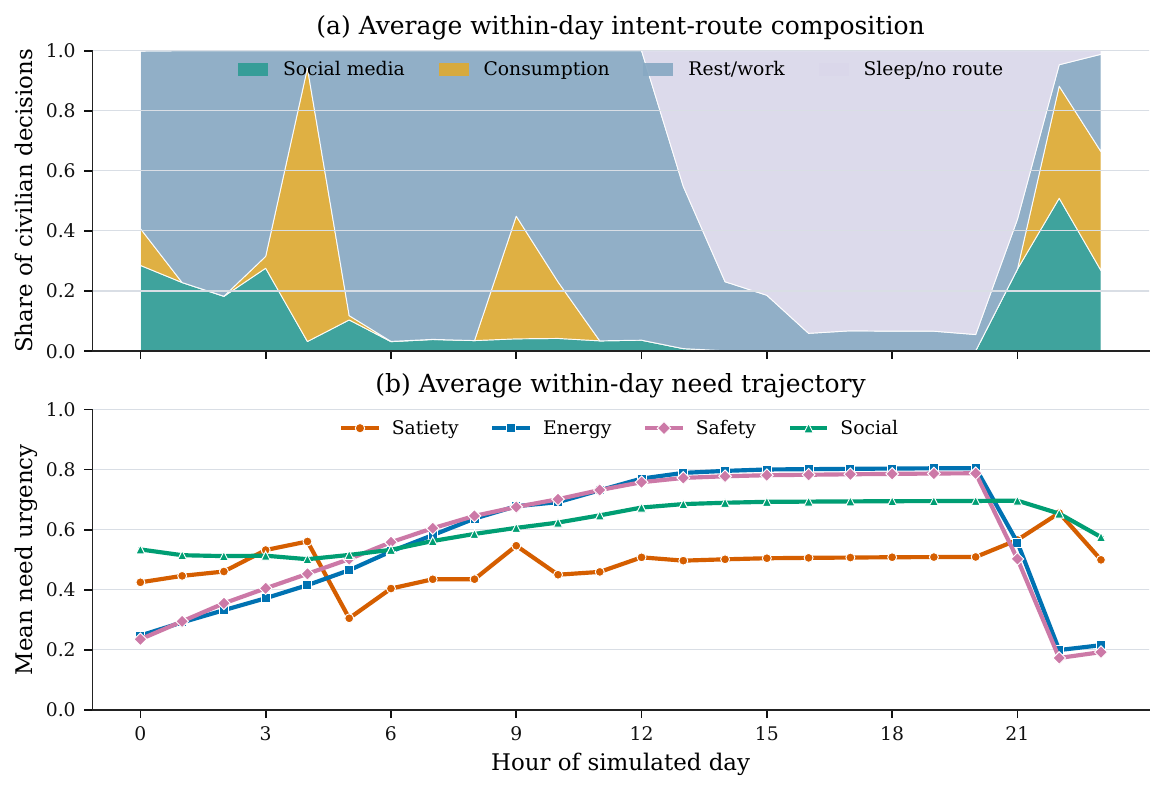}
\caption{Average within-day C0 need-intent dynamics. Panel (a) reports the mean hourly composition of civilian activity routes, averaged across all simulated days. Panel (b) reports the corresponding mean hourly trajectory of the four need-urgency variables.}
\label{fig:app-c0-diurnal-dynamics}
\end{figure*}

\paragraph{Average within-day dynamics.}
Figure~\ref{fig:app-c0-diurnal-dynamics} complements the day-level summary by averaging C0 behavior at each simulated hour across all 20 days and all civilians. The within-day route profile reveals the routine underlying the aggregate trend: rest/work dominates daytime hours, sleep/no-route dominates the late-day rest period, consumption is concentrated in short intervals, and social-media activity rises late in the day. The need trajectory exhibits a corresponding pattern, with energy, safety, and social urgency increasing during active hours and declining after the nighttime transition.

\section{Observation and State Schemas}
\label{app:schemas}

The civilian observation is assembled from several bounded views rather than represented as a single unstructured prompt:
\[
o_{i,t}
=
B_i\!\left(
p_i,\mathbf{s}_{i,t},e_{i,t},f_{i,t},d_{i,t},
\lambda_{i,t},g_{i,t},x_t
\right),
\]
where $p_i$ is the fixed profile, $e_{i,t}$ is the economic and mobility state, $f_{i,t}$ is the visible feed, $d_{i,t}$ is the bounded inbox and dialogue view, $\lambda_{i,t}$ is the civilian-visible campaign ledger, $g_{i,t}$ is local social context, and $x_t$ is the daily macroeconomic context. Before prompt construction, $B_i$ truncates each variable-length view (the public feed, for example, is limited to 10 posts), attaches stable identifiers required for execution grounding, and excludes latent campaign state that the civilian has not observed. Table~\ref{tab:app-observation-schema} summarizes the principal views available to civilian planning.

\begin{table*}[t]
\centering
\small
\setlength{\tabcolsep}{6pt}
\renewcommand{\arraystretch}{1.12}
\begin{tabular}{p{0.16\textwidth}p{0.28\textwidth}p{0.46\textwidth}}
\toprule
View & Main fields & Role in decision making \\
\midrule
Profile & age, gender, education, occupation, work hours, Big-Five traits & Conditions sleep, routine intent, spending style, and social response. \\
Need state & satiety, energy, safety, social, active need, previous intent & Provides the NIA salience signal for route selection. \\
Economic state & balance, monthly income, payday recency, recent purchases & Grounds affordability, scarcity pressure, and routine consumption. \\
Social feed & post id, author id, content, likes, comments, recent comments, campaign metadata & Exposes public campaign posts and peer or monitor warning comments. \\
Inbox & message id, message type, family, source role, source channel, sender id, content, contact count & Exposes private campaign contact, warning messages, and follow-up messages. \\
Dialogue state & whether the civilian replied, whether the sender replied last, turn count, recent messages & Distinguishes ongoing private conversations from independent single-turn messages. \\
Ledger view & total paid, returned amount, investment count, last payment, pending withdrawal, delay flags & Makes later decisions depend on prior payments, returns, and withdrawal friction. \\
Social context & trusted peers, community peers, peer investment signals & Allows forwarding, peer warning, and social proof effects. \\
Mobility/POI & current home/work area and nearby restaurant, cafe, and supermarket options & Grounds routine consumption alternatives to social-media attention. \\
\bottomrule
\end{tabular}
\caption{Civilian observation views assembled before intent and action planning.}
\label{tab:app-observation-schema}
\end{table*}

The campaign ledger records both campaign-level and victim-level pathway information. Table~\ref{tab:app-ledger-schema} lists the fields most relevant to fraud-pathway analysis. In the ledger-feedback ablation, the underlying ledger continues to record payments and withdrawals, whereas civilian-visible fields for cumulative payment, returned amount, investment count, last payment, and delay signals are masked or set to zero. This condition tests whether observable financial feedback is required to reproduce repeated-payment and withdrawal-friction dynamics.

For victim $v$ and campaign $c$, let $I_{v,c,t}$ denote cumulative gross fraud payments, $Y_{v,c,t}$ cumulative returns plus completed-withdrawal payouts, $N_{v,c,t}$ the payment count, and $W_{v,c,t}$ the withdrawal state. Given a grounded payment $P_{v,c,t}\geq0$ and payout $Q_{v,c,t}\geq0$, the accounting state is updated as
\[
\begin{aligned}
I_{v,c,t+1} &= I_{v,c,t}+P_{v,c,t},\\
Y_{v,c,t+1} &= Y_{v,c,t}+Q_{v,c,t},\\
N_{v,c,t+1} &= N_{v,c,t}+\mathbf{1}[P_{v,c,t}>0],\\
L^{\mathrm{net}}_{v,c,t+1}
&=\max\!\left(I_{v,c,t+1}-Y_{v,c,t+1},0\right).
\end{aligned}
\]
A withdrawal request updates $W_{v,c,t}$ without immediately changing $Y_{v,c,t}$; only an executed payout changes the cumulative returned amount. The append-only event record preserves message, campaign, agent, transaction, and hourly-step provenance, thereby distinguishing agent-visible observations from the global ledger used for execution and evaluation.

\begin{table*}[t]
\centering
\small
\setlength{\tabcolsep}{6pt}
\renewcommand{\arraystretch}{1.12}
\begin{tabular}{p{0.17\textwidth}p{0.32\textwidth}p{0.41\textwidth}}
\toprule
Schema group & Fields & Meaning \\
\midrule
Campaign identity & campaign id, fraud family, victim age bin, interaction channel & Links every loss to a campaign, FTC-aligned age bin, fraud family, and contact channel. \\
Contact path & first/last contact time, total contact count, received message identifiers & Reconstructs exposure and private-contact history before payment. \\
Payment ledger & total invested, total returned, net loss, investment count, last payment amount & Records gross payments, returns, current net loss, and repeated-payment depth. \\
Withdrawal ledger & withdrawal attempts, pending amount, delayed count, completed count & Records recovery attempts, delayed exits, and payout feedback. \\
Provenance & prior legal transaction count, social-proof exposure count, policy and ablation modes & Stores experimental and pathway metadata for later stratified analysis. \\
\bottomrule
\end{tabular}
\caption{Campaign ledger fields used for pathway reconstruction and aggregate metric computation.}
\label{tab:app-ledger-schema}
\end{table*}

\section{Evaluation Units and Reported Outcomes}
\label{app:outcomes}

The analysis uses two related but distinct accounting units. Let $I_{v,c}$ and $Y_{v,c}$ denote the final gross payment and returned amount, respectively, for victim $v$ in campaign $c$, and define
\[
L^{\mathrm{net}}_{v,c}=\max(I_{v,c}-Y_{v,c},0).
\]
Micro-level pathway and intervention outcomes use the set of positive-net-loss victim-campaign records
\[
\mathcal{R}^{+}
=
\{(v,c):L^{\mathrm{net}}_{v,c}>0\}.
\]
If $N_{v,c}$ is the number of payments and $A_{v,c}$ the number of withdrawal attempts in a record, the behavioral outcomes in the main paper are
\[
\begin{aligned}
\mathrm{TotalPositiveLoss}
&=\sum_{(v,c)\in\mathcal{R}^{+}}L^{\mathrm{net}}_{v,c},\\
\mathrm{PaymentCount}
&=\sum_{(v,c)\in\mathcal{R}^{+}}N_{v,c},\\
\mathrm{RepeatRate}
&=\frac{\sum_{(v,c)\in\mathcal{R}^{+}}\mathbf{1}[N_{v,c}\geq2]}
{|\mathcal{R}^{+}|},\\
\mathrm{WithdrawalRate}
&=\frac{\sum_{(v,c)\in\mathcal{R}^{+}}\mathbf{1}[A_{v,c}\geq1]}
{|\mathcal{R}^{+}|}.
\end{aligned}
\]
The loss-concentration curve similarly orders records in $\mathcal{R}^{+}$ by net loss; consequently, a civilian may contribute multiple campaign-specific pathways. Unique-victim statistics deduplicate records by $v$. By contrast, Quartet Distance aggregates gross payments once per unique civilian across campaigns, as formalized in Section~\ref{app:qd}. This distinction prevents simulation-specific returns from altering the regulator-grounded QD assessment while retaining returns and withdrawals as behavioral outcomes.

\begin{table*}[tb]
\centering
\footnotesize
\setlength{\tabcolsep}{4pt}
\renewcommand{\arraystretch}{1.12}
\begin{tabular*}{\textwidth}{@{\extracolsep{\fill}}lrrrr}
\toprule
Condition & \makecell[r]{Total positive\\loss (\$)} & \makecell[r]{Payment\\count} & \makecell[r]{Repeat-payment\\rate} & \makecell[r]{Withdrawal-attempt\\rate} \\
\midrule
C0 No warning & 144,999.50 & 623 & 49.4\% & 8.6\% \\
C1 Platform broadcast & 96,170.10 & 205 & 38.2\% & 17.1\% \\
C2 Monitor labels & 70,220.55 & 284 & 41.5\% & 22.6\% \\
C3 Random comments & 42,014.25 & 141 & 49.1\% & 18.2\% \\
C4 Peer warnings & 24,752.50 & 155 & 43.1\% & 21.6\% \\
C5 Oracle personalized & 17,916.00 & 117 & 10.9\% & 47.5\% \\
\bottomrule
\end{tabular*}
\caption{Numerical values corresponding to the counter-fraud panel in the main paper. The final protocol assigns every condition a seed-42 run over 480 hours beginning at 08:00 on March 11, 2025. Rate denominators comprise positive-net-loss victim--campaign records.}
\label{tab:app-intervention-results}
\end{table*}

\paragraph{Scale effects.}
Relative to C0, positive net loss decreases by 33.7\%, 51.6\%, 71.0\%,
82.9\%, and 87.6\% under C1--C5, respectively, while payment count decreases
by 67.1\%, 54.4\%, 77.4\%, 75.1\%, and 81.2\%.  The two sequences do not have
the same ordering.  C3 produces only 141 payments, fewer than the 155 under C4,
yet its \$42,014.25 positive net loss is 69.7\% higher; equivalently, C4 lowers
loss by a further 41.1\% despite recording slightly more payments.  Payment
count therefore measures activity frequency but does not by itself identify
the monetary severity of the surviving pathways.

\paragraph{Frequency--severity decomposition.}
To separate these dimensions, define the descriptive loss intensity
\[
\bar\ell_c=
\frac{\mathrm{TotalPositiveLoss}_c}{\mathrm{PaymentCount}_c}.
\]
This is not mean transfer size: the numerator is positive net loss after
returns, whereas the denominator counts payments in the corresponding
positive-net-loss records.  It only measures the final net-loss burden per
counted payment.  The value is \$232.74 under C0 and \$469.12, \$247.26,
\$297.97, \$159.69, and \$153.13 under C1--C5.  C1 sharply reduces frequency,
but its remaining loss intensity is 2.02 times the C0 value, indicating that
after broad broadcasts remove many low-loss or low-frequency paths, the
remaining positive-loss sample is still dominated by higher-value paths.  This
is a conditional-sample composition effect, not evidence that broadcasting
causes larger individual losses.  C2 remains close to C0 on intensity.  C3 has
the smallest non-oracle payment count but remains above C0 on intensity,
whereas C4 reduces intensity to 68.6\% of C0.  C4's comparative advantage is
therefore more consistent with compressing high-value escalation than merely
removing events.

\paragraph{Escalation and exit.}
Repeat-payment and withdrawal-attempt rates further distinguish entry
prevention, escalation control, and exit seeking.  C3's 49.1\% repeat rate is
nearly identical to C0's 49.4\%; together with its lower payment count, this is
consistent with random comments preventing some paths from entering while
providing little persistent constraint on the remaining paid paths.  C1 has
the lowest non-oracle repeat rate (38.2\%) and raises withdrawal attempts from
8.6\% to 17.1\%, but its high loss intensity shows that broad coverage does not
fully reach the remaining severe paths.  C4 does not minimize repeat payment
(43.1\%), yet combines the lowest non-oracle loss intensity with a 21.6\%
withdrawal-attempt rate.  Its principal gain is thus closer to limiting
high-value escalation and encouraging some exits than to eliminating every
repeat payment.  C5 combines a 10.9\% repeat rate, a 47.5\% withdrawal-attempt
rate, and \$153.13 loss intensity, giving the clearest stop-additional-payment
and seek-exit signature.  Because it uses transaction truth and
campaign-specific context, it remains an information-rich timing upper bound.

\paragraph{Interpretive boundary.}
The rate denominator in each condition is the set of victim--campaign records
that still have positive net loss, not a common civilian cohort.  An
intervention changes both membership in this denominator and subsequent path
length, so cross-condition rate differences combine behavioral and
compositional change and are not individual treatment effects.  The conditions
also jointly change trigger timing, delivery channel, and information
specificity.  Finally, the table reports complete runs for one fixed profile
realization and seed, without cross-seed variance or confidence intervals.
The unified protocol controls the start clock and payday timing across C0--C5,
but it does not by itself identify channel-specific causal effects.  These
contrasts support trajectory-consistent mechanism hypotheses rather than
statistical identification of such effects.

\section{Victimization Pathway Diagnostics}
\label{app:pathway-diagnostics}

We reconstruct micro-level pathways from the completed common-clock C0 run using the positive-net-loss victim--campaign unit in Section~\ref{app:outcomes}. The run contains 81 such records associated with 63 unique civilians and \$144,999.50 in total positive net loss. All 63 civilians received campaign contact and a private message before their first fraud payment. The median first-contact-to-first-payment lag is 72 simulated hours. The seven pre-payment contacts reported in the main paper are counted from timestamp-filtered campaign-contact events before the first transfer; the exported \texttt{total\_contact\_count} is a final-horizon field and has median 16. The latter must not be interpreted as a pre-payment exposure count.

\begin{figure*}[tb]
\centering
\includegraphics[width=0.97\textwidth]{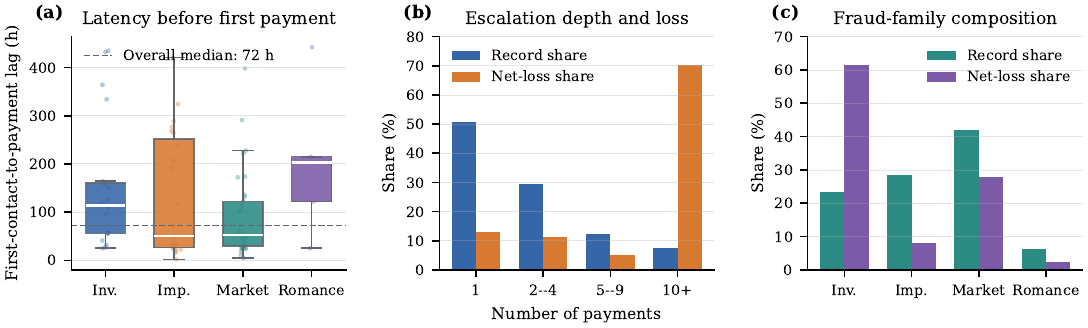}
\caption{C0 victimization-pathway diagnostics. Panel (a) shows the distribution of first-contact-to-first-payment latency for positive-net-loss victim--campaign records; points are individual records, boxes show the interquartile range, and the dashed line marks the overall 72-hour median. Panel (b) contrasts the prevalence of each final payment-depth group with its share of positive net loss. Panel (c) contrasts each fraud family's record share with its net-loss contribution. Record share weights every victim--campaign record equally, whereas net-loss share weights records by their positive net-loss amounts.}
\label{fig:app-pathway-diagnostics}
\end{figure*}

Figure~\ref{fig:app-pathway-diagnostics} exposes three complementary pathway
properties.  Panel (a) shows median conversion lags of 50 and 52 hours for
impersonation and marketplace fraud, 114 hours for investment fraud, and 203
hours for romance fraud, against the 72-hour overall median.  Panel (b)
partitions final payment count into the mutually exclusive groups
$S\in\{1,2\text{--}4,5\text{--}9,10+\}$.  For each group, record share is
$|S|/|\mathcal R^+|$, whereas net-loss share is
$\sum_{(v,c)\in S}L^{\mathrm{net}}_{v,c}/144{,}999.50$.  Combining the last
three groups, 40 records (49.4\%) contain at least two payments but account for
86.8\% of total positive loss.  Combining the last two groups, 16 records
(19.8\%) contain at least five payments and account for 75.6\%, while the six
records (7.4\%) in the 10+ group account for 70.4\%.  Panel (c)
applies the same count- versus amount-weighted comparison to fraud families:
investment fraud represents only 23.5\% of records but contributes 61.6\% of
net loss.  The persistent gap between the two shares demonstrates that
financial harm is concentrated in a small set of high-value, deeply escalated
paths.  Sorting all records by net loss and taking
$\lceil0.1\times81\rceil=9$ records yields a top-decile share of 81.1\%.

\begin{table*}[t]
\centering
\footnotesize
\setlength{\tabcolsep}{3pt}
\renewcommand{\arraystretch}{1.12}
\begin{tabular*}{\textwidth}{@{\extracolsep{\fill}}lrrrrrr}
\toprule
Fraud family & Records & \makecell[r]{Unique\\victims} & \makecell[r]{Net loss\\(\$)} & \makecell[r]{Loss\\share} & \makecell[r]{Median\\lag (h)} & \makecell[r]{Median final\\contacts} \\
\midrule
Investment & 19 & 18 & 89,350.00 & 61.6\% & 114 & 16 \\
Impersonation & 23 & 20 & 11,590.00 & 8.0\% & 50 & 12 \\
Marketplace & 34 & 31 & 40,459.50 & 27.9\% & 52 & 18.5 \\
Romance & 5 & 5 & 3,600.00 & 2.5\% & 203 & 29 \\
\midrule
All records & 81 & 63 & 144,999.50 & 100.0\% & 72 & 16 \\
\bottomrule
\end{tabular*}
\caption{Fraud-family decomposition of positive-net-loss C0 pathways.
``Unique victims'' is deduplicated within each row; because a civilian may
appear in more than one campaign, family-specific unique-victim counts do not
sum to the all-record value.  Contact counts are final-horizon ledger fields.}
\label{tab:app-pathway-families}
\end{table*}

The family decomposition also reveals different temporal regimes.  Investment
records contribute 61.6\% of positive net loss despite representing 23.5\% of
records.  Marketplace pathways are more frequent and have a median of two
payments, whereas the median payment count in each other family is one.
Romance pathways are few but have the longest median contact-to-payment delay
(203 hours) and the largest median final contact count (29).  These statistics
are descriptive properties of one fixed-seed trajectory rather than
family-level causal estimates, but they demonstrate that the aggregate loss
surface is supported by heterogeneous multi-step pathways.

\begin{minipage}{\textwidth}
\centering
\footnotesize
\setlength{\tabcolsep}{5pt}
\renewcommand{\arraystretch}{1.14}
\begin{tabular}{p{0.12\textwidth}p{0.37\textwidth}p{0.39\textwidth}}
\toprule
Simulation time & Information state and recorded reasoning & Financial state transition \\
\midrule
Mar.\ 15, 05:00--Mar.\ 17, 04:00 & The civilian repeatedly asks about task content, the refund process, and the minimum advance. The operator describes phone-based tasks, an advance range of \$52--\$315, payment through common mobile channels, and a return ``generally within one hour.'' & No fraud payment is executed during the 48-hour cultivation interval. \\
Mar.\ 17, 05:00 & The operator lowers the actionable threshold to \$52. The civilian records that the amount is low enough to try. & First advance: \$52. \\
Mar.\ 17, 08:00 & A \$15 ``first-task profit'' notification is delivered together with a new-task invitation. & The ledger records a \$15 return, reducing the contemporaneous net exposure to \$37. \\
Mar.\ 17, 09:00--12:00 & Repeated notifications couple the realized small return with new tasks capped at \$100. Before the fourth transfer, the civilian explicitly records that they had asked repeatedly and ``received returns.'' & Three further \$100 advances are executed at 09:00, 11:00, and 12:00. Gross payment reaches \$352 within seven hours of entry. \\
Mar.\ 17, 13:00 & The civilian records that losses remain substantial despite returns and asks to reduce risk. & A \$100 withdrawal is requested; no withdrawal payout is completed. \\
Mar.\ 17, 20:00--Mar.\ 18, 08:00 & Additional small return notices continue while the withdrawal remains unresolved. & Four later returns total \$60.60. Across the full path, returns total \$105.60. \\
End of run & The campaign continues to present task and refund assurances, but no fifth payment occurs. & Final campaign state: four payments, \$352 gross paid, \$105.60 returned, \$246.40 net loss, 14 withdrawal delays, and \$100 still pending. \\
\bottomrule
\end{tabular}
\captionof{table}{A provenance-linked C0 marketplace-fraud pathway. Information events and financial events are shown in separate columns so that persuasive content, the civilian's stated reason, and the executed ledger transition are not conflated.}
\label{tab:app-c0-case}
\end{minipage}
\vspace{0.8em}

\section{Mechanism Case Studies}
\label{app:case-studies}

\paragraph{Scope and selection.}
The cases in this section are reconstructed from the completed common-clock C0--C5 runs and retain their recorded simulated timestamps. They are diagnostic mechanism traces drawn from the same fixed-seed executions summarized above, not additional observations, representative-sample estimates, or individual counterfactuals. A C0 pathway was selected only if the run artifacts linked visible campaign information, grounded transfers, returns, and withdrawal state. The selected marketplace-fraud record is mechanistically representative rather than statistically average: its 48-hour contact-to-payment lag is close to the 52-hour marketplace median, its four payments lie in the common two-to-four-payment group, and it contains both positive reinforcement and an attempted exit. For C1--C5, a case was retained only when the intervention was visible to the civilian, the civilian's recorded reason explicitly responded to it, and the subsequent financial trace was consistent with entry inhibition, escalation control, or exit seeking.

\subsection{C0: Information--Financial Escalation in One Marketplace Path}

Table~\ref{tab:app-c0-case} joins the information and financial streams for civilian V82 and the ``Yuanyu Part-time Platform'' campaign. The reconstruction uses campaign identifiers and simulated timestamps rather than row adjacency: a message is treated as a financial trigger only when the executed transfer carries the corresponding information provenance.

\paragraph{Mechanism interpretation.}
This path exhibits threshold adaptation, reinforcement, rapid escalation, and exit friction in a compact sequence. The initial broad advance range does not immediately produce payment; conversion occurs only after the operator offers its \$52 lower bound. The first realized \$15 return then becomes both a financial event and new persuasive information. Within four hours, the civilian makes three additional payments, and the final payment reason explicitly invokes prior returns. The subsequent withdrawal request shows that positive reinforcement did not eliminate risk perception, but the request also did not restore funds: only small operator-controlled returns were executed, while the requested \$100 remained pending. Consequently, ``returned amount'' and ``successful exit'' are empirically distinct states. The case also illustrates the model's closed loop: information changes payment behavior, ledger outcomes are surfaced as later information, and that information conditions both escalation and withdrawal.

\subsection{Intervention-Consistent Micro-Cases}

Table~\ref{tab:app-intervention-cases} reports the four C1--C5 conditions for which the completed run trace satisfies the stated inclusion rule. The table deliberately omits a claimed C3 individual effect. C3 does contain the injected generic comments, but the run artifacts do not provide a single civilian trace that simultaneously links one such comment to an explicit response reason and a subsequent financial inhibition outcome. This absence of a retained case is an evidence-provenance limitation, not evidence that C3 has zero aggregate effect.

\begin{table*}[tb]
\centering
\footnotesize
\setlength{\tabcolsep}{4.5pt}
\renewcommand{\arraystretch}{1.14}
\begin{tabular}{p{0.09\textwidth}p{0.23\textwidth}p{0.325\textwidth}p{0.255\textwidth}}
\toprule
Condition & Pre-intervention path and visible warning & Civilian response recorded in the trace & Observed financial implication \\
\midrule
C1 & V13 repeatedly inquires about a part-time-work campaign, including whether a \$50 trial is available. At hour 72, the fixed platform DM warns against transfers, deposits, unlocking fees, and guaranteed returns. & The same hourly decision records: ``received a safety reminder and decided not to process the part-time-work invitation.'' The campaign dialogue ends at that decision. & No payment to that campaign, or to any fraud campaign, is recorded over the 480-hour run. This is consistent with entry inhibition, although it is not an individual counterfactual estimate. \\
C2 & V75 repeatedly sees monitor-annotated investment posts. The appended label states that the content may involve fraud and advises against blind transfers. & Across hours 9--34, the reasons repeatedly state that the system marked the post suspicious and that the civilian would ``observe first'' or ``handle it cautiously.'' Actions remain low-cost likes rather than private inquiry. & No fraud payment is recorded for V75 over the full horizon. The trace is consistent with the label constraining escalation from public attention to private contact and payment. \\
C4 & V12 pays \$500 once to an investment campaign. The civilian subsequently produces peer warnings based on the loss and observes an environment in which the platform is repeatedly reported. & At hour 419, the reason states: ``the platform has been repeatedly reported and is risky; request withdrawal to reduce losses.'' & The record remains at one payment and adds a \$500 withdrawal request. No payout is completed, so C4 inhibits further escalation and induces exit seeking without guaranteeing recovery. \\
C5 & V5 pays a \$100 verification deposit. The next-hour oracle DM identifies the named platform and amount, instructs the civilian to stop transferring, withdraw, preserve evidence, and verify through official channels. & In that hour, the civilian records that the anti-fraud reminder made the platform appear fraudulent, requests withdrawal, and marks the campaign suspicious. & The path stops after one payment and carries a \$100 pending withdrawal. This is the clearest transaction-to-warning-to-exit chain, while its use of simulator truth confirms that C5 is an idealized upper bound. \\
\bottomrule
\end{tabular}
\caption{Intervention-consistent micro-cases from the completed common-clock runs. ``Consistent with'' is used deliberately: these within-run traces expose plausible mechanisms but do not supply untreated counterfactuals for the same civilian.}
\label{tab:app-intervention-cases}
\end{table*}

\paragraph{Cross-case reading.}
The four cases separate three forms of suppression that the aggregate rates combine. C1 interrupts an active dialogue before entry; C2 keeps exposure at a low-cost public-interaction stage; and C4 and C5 act after payment by limiting repetition and inducing withdrawal. Warning specificity and timing increase across these examples. The generic C1 message can terminate a matched dialogue but does not identify a particular transaction, while the C2 label is content-specific but competes within the feed. C4 carries socially generated, platform-specific evidence but emerges endogenously and may arrive after harm. C5 reacts immediately with exact transaction context, which explains the unusually direct reasoning response but also makes the mechanism non-deployable without an external detector that can approximate the oracle signal. None of the exit-seeking cases completes recovery, underscoring that suppressing additional payment and reversing an existing loss are separate outcomes.

\FloatBarrier
\section{Quartet Distance Details}
\label{app:qd}

Quartet Distance evaluates the simulated distribution of fraud losses against four FTC-grounded empirical regularities. All four QD components are computed from gross fraud payments. Returns and withdrawal payouts are not deducted because the FTC targets characterize reported fraud-loss severity rather than simulation-specific return and withdrawal processes.

Let $V^+$ be the set of unique civilians with positive gross fraud payment, and let
\[
L_v=\sum_c I_{v,c}
\]
be the gross payment of civilian $v$, aggregated across campaigns. Let $B$ denote the seven adult FTC age bins:
\[
B=\{20\text{--}29,30\text{--}39,40\text{--}49,50\text{--}59,
60\text{--}69,70\text{--}79,\text{80+}\}.
\]

\paragraph{HTL: relative high-loss tail.}
The FTC high-loss threshold is $\$10{,}000$, while the FTC overall median positive-loss amount is $\$497$~\citep{ftc2025databook}. Accordingly, the high-loss threshold is converted to the relative multiplier
\[
\lambda = 10000/497 = 20.12.
\]
For a simulation run, the high-loss threshold is $\tau=\lambda \cdot \mathrm{median}_{v\in V^+}(L_v)$. The simulated high-loss share is
\[
\hat{p}_{\mathrm{HTL}}=
\frac{1}{|V^+|}\sum_{v\in V^+}\mathbf{1}[L_v>\tau].
\]
The target is $p_{\mathrm{HTL}}=0.126$, computed from the FTC 2024 Data Book category for reports exceeding $\$10{,}000$. The distance is
\[
d_{\mathrm{HTL}}=
\frac{|\hat{p}_{\mathrm{HTL}}-0.126|}{0.126}.
\]
This relative error is not clipped. Because
$\hat{p}_{\mathrm{HTL}}\in[0,1]$, its implementation range is
$0\leq d_{\mathrm{HTL}}\leq(1-0.126)/0.126\approx6.94$; values greater
than one are therefore admissible.

\paragraph{ALS: all-bin age-loss severity profile.}
For each age bin $b$, let $\hat{m}_b$ be the simulated median of $L_v$ over positive-loss civilians in that bin. The target medians $m_b$ are the FTC age-bin medians listed in Table~\ref{tab:app-ftc-age-targets}. The all-bin profile distance is
\[
d_{\mathrm{ALS}}=
\frac{1}{|B|}\sum_{b\in B}
\min\left(\frac{|\hat{m}_b-m_b|}{\max(m_b,1)},2\right).
\]
The upper cap limits the contribution of an age bin with a highly discrepant or small simulated median. It bounds the component at two, rather than normalizing it to the unit interval: $d_{\mathrm{ALS}}\in[0,2]$.

\paragraph{FSH: fraud-family severity hierarchy.}
For fraud-family severity, let $f_c$ denote the fraud family of campaign $c$. The simulator compares the median gross victim--campaign payment for investment fraud with the corresponding median for impersonation fraud:
\[
\hat{r}_{\mathrm{FSH}}=
\frac{\operatorname{median}\{I_{v,c}:f_c=\mathrm{investment}\}}
{\max\!\left(
\operatorname{median}\{I_{v,c}:f_c=\mathrm{impersonation}\},1
\right)}.
\]
The FTC target ratio is $r_{\mathrm{FSH}}=9196/800=11.50$, using the reported median losses for the \emph{Investment Related} and \emph{Imposter Scams} categories. The log-scale distance is
\[
d_{\mathrm{FSH}}=
\frac{|\log(\hat{r}_{\mathrm{FSH}}/11.50)|}{|\log(11.50)|}.
\]
For a valid positive $\hat{r}_{\mathrm{FSH}}$, this log-ratio discrepancy is not clipped and has no finite upper bound. If the ratio cannot be formed or is nonpositive, the implementation assigns the fallback value $d_{\mathrm{FSH}}=1$.

\paragraph{ALR: age loss-report share.}
FTC age-specific positive-loss report shares are estimated from each adult age bin's fraud-report count $N_b$ and percentage reporting a fraud loss $q_b$. The target share is
\[
a_b=\frac{N_bq_b}{\sum_{b'\in B}N_{b'}q_{b'}}.
\]
The simulated share $\hat{a}_b$ is the share of unique positive-loss civilians in age bin $b$. The distance is the total variation distance:
\[
d_{\mathrm{ALR}}=
\frac{1}{2}\sum_{b\in B}|\hat{a}_b-a_b|.
\]
As a total variation distance between age-share distributions, this component satisfies $d_{\mathrm{ALR}}\in[0,1]$.

\paragraph{Composite.}
The age-aware QD used in the paper is
\[
\mathrm{QD}=
\frac{
d_{\mathrm{HTL}}+
d_{\mathrm{ALS}}+
d_{\mathrm{FSH}}+
d_{\mathrm{ALR}}
}{4}.
\]
No additional clipping or rescaling is applied before aggregation. Consequently, $\mathrm{QD}$ is a nonnegative, lower-is-better discrepancy measure, not a probability or a score normalized to $[0,1]$. It may exceed one; moreover, the HTL and FSH components may individually exceed one and can dominate the arithmetic mean. Values should therefore be interpreted together with the four reported components.

\begin{table}[t]
\centering
\footnotesize
\setlength{\tabcolsep}{3pt}
\renewcommand{\arraystretch}{1.05}
\begin{tabular}{@{}L{0.23\columnwidth}L{0.20\columnwidth}L{0.49\columnwidth}@{}}
\toprule
Quantity & Range & Scale-defining operation \\
\midrule
$d_{\mathrm{HTL}}$ & $[0,6.94]$ & Unclipped relative error against the target tail share. \\
$d_{\mathrm{ALS}}$ & $[0,2]$ & Mean seven-bin relative error, capped at two per bin. \\
$d_{\mathrm{FSH}}$ & $[0,\infty)$ & Unclipped normalized log-ratio error; fallback value one when unavailable. \\
$d_{\mathrm{ALR}}$ & $[0,1]$ & Total variation distance between age-share distributions. \\
$\mathrm{QD}$ & $[0,\infty)$ & Unclipped arithmetic mean of the four component distances. \\
\bottomrule
\end{tabular}
\caption{Ranges of the QD component and composite distances under the current implementation. The quantities do not share a common unit-interval normalization.}
\label{tab:app-qd-ranges}
\end{table}

\begin{table}[t]
\centering
\scriptsize
\setlength{\tabcolsep}{2.5pt}
\renewcommand{\arraystretch}{1.0}
\begin{tabular*}{\columnwidth}{@{\extracolsep{\fill}}lrrrr}
\toprule
Age & Median & $N_b$ & $q_b$ & $a_b$ \\
\midrule
20--29 & 417 & 155,346 & 0.44 & 0.1735 \\
30--39 & 450 & 203,764 & 0.40 & 0.2069 \\
40--49 & 500 & 191,508 & 0.38 & 0.1847 \\
50--59 & 520 & 175,898 & 0.35 & 0.1563 \\
60--69 & 691 & 208,896 & 0.29 & 0.1538 \\
70--79 & 1,000 & 159,550 & 0.24 & 0.0972 \\
80+ & 1,650 & 51,713 & 0.21 & 0.0276 \\
\bottomrule
\end{tabular*}
\caption{FTC age-bin constants used by the ALS and ALR components. The 19-and-under bin is omitted because the simulated population contains no minors.}
\label{tab:app-ftc-age-targets}
\end{table}

\paragraph{Raw diagnostic decomposition.}
The completed common-clock full-model run contains 51 unique civilians with
positive gross payment and 65 positive victim--campaign payment records.  Its
median unique-victim gross payment is \$200, giving a relative HTL threshold
of \$4,024.14.  Seven of the 51 civilians exceed this threshold, so the
simulated tail share is $7/51=0.1373$ and
$d_{\mathrm{HTL}}=0.0893$.  The investment and impersonation
victim--campaign medians are \$3,000 and \$125, respectively, producing
$\hat r_{\mathrm{FSH}}=24$ and $d_{\mathrm{FSH}}=0.3012$.
The complete component decomposition is shown in
Table~\ref{tab:app-qd-raw-diagnostics}.

\begin{table*}[t]
\centering
\small
\setlength{\tabcolsep}{7pt}
\renewcommand{\arraystretch}{1.12}
\begin{tabular*}{0.94\textwidth}{@{\extracolsep{\fill}}lp{0.39\textwidth}rr}
\toprule
Component & Simulation statistic & Target & Distance \\
\midrule
HTL & tail share $0.1373$ above \$4,024.14; median gross payment \$200 & 0.126 & 0.0893 \\
ALS & seven-bin simulated medians shown in Figure~\ref{fig:app-qd-diagnostics}(a) & FTC median profile & 0.6513 \\
FSH & investment-to-impersonation median gross-payment ratio $3000/125=24.0$ & 11.50 & 0.3012 \\
ALR & seven-bin simulated unique-victim shares shown in Figure~\ref{fig:app-qd-diagnostics}(b) & FTC loss-report profile & 0.1495 \\
\midrule
$\mathrm{QD}$ & arithmetic mean of the four distances, without clipping & 0 & 0.2978 \\
\bottomrule
\end{tabular*}
\caption{Raw-to-distance audit trail for the full-model QD value in the main
paper.  The zero in the composite target column denotes exact agreement, not
an empirical zero-loss target.}
\label{tab:app-qd-raw-diagnostics}
\end{table*}

\begin{figure*}[t]
\centering
\includegraphics[width=0.97\textwidth]{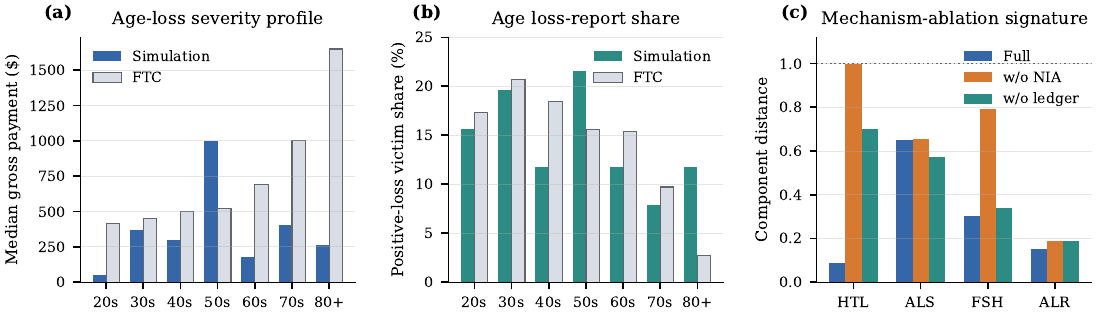}
\caption{QD diagnostics.  Panels (a) and (b) compare the full model with the
FTC targets used by ALS and ALR.  Panel (c) decomposes the full model and the
two mechanism ablations.  The dotted line at one is an interpretive reference,
not a normalization boundary; component distances are permitted to exceed it.}
\label{fig:app-qd-diagnostics}
\end{figure*}

\begin{table*}[!t]
\centering
\small
\setlength{\tabcolsep}{5pt}
\renewcommand{\arraystretch}{1.12}
\begin{tabular}{p{0.16\textwidth}p{0.28\textwidth}p{0.19\textwidth}p{0.27\textwidth}}
\toprule
System & Loss adapter & Age and family support & Composite convention in the main-paper table \\
\midrule
TransXion & Sum flagged transaction amounts by source account; join \texttt{person\_age} and retain positive aggregates & age available; no compatible individual-targeted fraud-family label & FSH is reported as unavailable.  The displayed 0.4894 is the mean of the three available distances, $(0.3722+0.9168+0.1793)/3$, and is therefore a partial composite. \\
MultiAgent4\-Fraud & Aggregate successful victim-originating transfers to malicious agents by civilian profile in the archived evaluation output & profile age and assigned malicious-agent fraud category available to the audit adapter & Mean of all four reported distances: $(0.6211+0.6586+0.6017+0.5243)/4=0.6014$. \\
\textsc{FraudSociety} & Sum gross payments across campaigns once per civilian for HTL, ALS, and ALR; retain victim--campaign family records for FSH & explicit age, campaign identity, and fraud family & Four-component QD: $(0.0893+0.6513+0.3012+0.1495)/4=0.2978$. \\
\bottomrule
\end{tabular}
\caption{Source-to-QD alignment used for the baseline comparison.  A partial
composite is not numerically equivalent to strict four-component QD; the dash
for TransXion's FSH in the main-paper table records this limitation.}
\label{tab:app-baseline-alignment}
\end{table*}

\section{External-Baseline Alignment Protocol}
\label{app:baseline-alignment}

QD postprocessing is applied through source-specific adapters.  The adapters
do not assume that native records from transaction simulators, social
simulators, and \textsc{FraudSociety} are semantically identical.  They first
identify the finest supported positive-loss unit, aggregate gross amounts at
that unit, join profile age where available, and map a fraud family only when
the source explicitly provides one.  The four formulas and FTC constants in
Section~\ref{app:qd} are then applied without dataset-specific tuning,
min--max scaling, or imputation of a missing component.

For TransXion, the adapter uses \texttt{tx.csv} to select flagged records and
aggregate \texttt{Amount Paid} by \texttt{From Account}, then joins
\texttt{person.csv} for age binning.  This account-level proxy audits a
relative transaction-loss tail and an age profile, but it is not an
individual-targeted victim--campaign ledger.  MultiAgent4Fraud provides
profile-grounded social interaction and terminal transfer events; its adapter
treats a successful transfer as gross victim loss and does not manufacture
post-transfer returns, withdrawals, or repeated-payment states that the source
system does not simulate.  These transformations make the displayed
components reproducible while preserving a clear scope limitation: the
comparison evaluates compatible output statistics, not equality of native
mechanisms or event semantics.

\FloatBarrier
\section{Reproducibility Specification and Artifacts}
\label{app:reproducibility}

Table~\ref{tab:app-runtime-configuration} records the configuration of the
120-agent, 20-day experiments.  In particular, the canonical entry point defaults to 168 steps,
so reproducing the reported horizon requires explicitly setting
\texttt{SIM\_NUM\_STEPS=480}.  The profile list is truncated deterministically
to the first 112 civilians while preserving all eight operators.  Every final
intervention and QD run uses the common start time
\texttt{2025-03-11T08:00:00}.

\begin{table}[tbp]
\centering
\small
\setlength{\tabcolsep}{5pt}
\renewcommand{\arraystretch}{1.12}
\begin{tabular}{p{0.20\textwidth}p{0.25\textwidth}p{0.45\textwidth}}
\toprule
Configuration group & Reported value & Implementation detail \\
\midrule
Clock and horizon & 3,600 s/tick; 480 ticks; March 11, 2025 at 08:00 & identical 20-day calendar for every condition; daily and payday triggers use the simulated date and hour \\
Population & 112 civilians, 8 operators, 8 campaigns & first 112 civilians in the fixed profile file; two operators/campaigns per fraud family \\
Randomness & simulator seed 42 & fixes profile truncation, graph sampling, and random-warning sampling; it does not guarantee bitwise-identical remote inference \\
Social bounds & feed 10 posts; inbox 20 unread DMs & observations carry stable post, message, author, sender, campaign, and transaction identifiers for execution grounding \\
Initial graph & geographic $k=6$; rewire $p=0.15$; operator degree-weighted targets $\leq10$ & relation labels and homophily/attachment additions are retained in the shared graph \\
LLM and sampling & Qwen3-14B; temperature 0 & validated JSON; recoverable syntax repair; invalid names, targets, and arguments rejected \\
Victim planning limits & step plan 256 tokens; fraud realization 256; ambient/night check 96 & principal concurrency limits are 12, 8, and 30/16, respectively; generic transport concurrency is 10 \\
Operator planning limits & phase 1,024 tokens; finance 768; social 768 & concurrency limits are 4, 2, and 4; timeouts are 30, 25, and 25 s, respectively \\
\bottomrule
\end{tabular}
\caption{Runtime configuration for the reported 120-agent experiments.  Token
and concurrency entries document request scheduling and output bounds rather
than behavioral parameters.}
\label{tab:app-runtime-configuration}
\end{table}

\subsection*{Unified-Clock Run Provenance and Result Audit}

The final tables use one completed seed-42 execution per row and do not pool
repeated executions.  Table~\ref{tab:app-run-provenance} summarizes the runs
completed under the common March 11 clock.  The reported intervention,
pathway, and QD analyses are computed from these runs; run identities and
content digests were checked during analysis before the values were reported.

\begin{table}[tbp]
\centering
\scriptsize
\setlength{\tabcolsep}{4pt}
\renewcommand{\arraystretch}{1.08}
\begin{tabular}{p{0.10\textwidth}p{0.46\textwidth}p{0.15\textwidth}p{0.20\textwidth}}
\toprule
Result & Condition-specific setting & Unified start & Run status \\
\midrule
C0 & all C1--C5 warning switches disabled & Mar.\ 11, 08:00 & completed \\
C1 & \texttt{USE\_PLATFORM\_WARNING=1} & Mar.\ 11, 08:00 & completed \\
C2 & \texttt{USE\_MONITOR\_AGENT=1} & Mar.\ 11, 08:00 & completed \\
C3 & \texttt{USE\_RANDOM\_WARNING=1} & Mar.\ 11, 08:00 & completed \\
C4 & \texttt{USE\_PEER\_WARNING=1} & Mar.\ 11, 08:00 & completed \\
C5 & \texttt{USE\_C5\_ORACLE\_WARNING=1} & Mar.\ 11, 08:00 & completed \\
\addlinespace
QD: full & full NIA and surfaced-ledger configuration & Mar.\ 11, 08:00 & completed \\
QD: w/o NIA & \texttt{USE\_NEED\_CONDITIONED\_INTENT=0} & Mar.\ 11, 08:00 & completed \\
QD: w/o ledger & \texttt{USE\_SURFACED\_LEDGER=0} & Mar.\ 11, 08:00 & completed \\
\bottomrule
\end{tabular}
\caption{Completed unified-clock runs used for the intervention, pathway, and
QD results.  Each row represents one seed-42 execution over the reported
480-hour horizon.}
\label{tab:app-run-provenance}
\end{table}

Before reporting, the completed run manifest was checked against artifact
digests, horizons, populations, the common start time,
Table~\ref{tab:app-intervention-results}, the four QD components, and their
composite.  Any row whose start time differed from
\texttt{2025-03-11T08:00:00} was rejected.

The following command is the baseline configuration template.  Each
intervention condition changes exactly one of the C1--C5 switches documented
in Table~\ref{tab:app-intervention-implementation}; the full-model ablations
instead change \texttt{USE\_NEED\_CONDITIONED\_INTENT} or
\texttt{USE\_SURFACED\_LEDGER}.  Output directories must be distinct because
all artifacts are run-local.

\begin{lstlisting}[language=bash,numbers=none,basicstyle=\scriptsize\ttfamily,
  frame=single,xleftmargin=0.5em,framexleftmargin=0.4em]
SIM_OUTPUT_DIR=runs/<run_name> \
SIM_NUM_STEPS=480 SIM_PROFILE_LIMIT=120 \
SIM_RANDOM_SEED=42 \
SIM_START_DATE=2025-03-11T08:00:00 \
USE_NEED_CONDITIONED_INTENT=1 \
USE_SURFACED_LEDGER=1 \
USE_PEER_WARNING=0 USE_PLATFORM_WARNING=0 \
USE_MONITOR_AGENT=0 USE_RANDOM_WARNING=0 \
USE_C5_ORACLE_WARNING=0 \
python -u env_main_fraud_simulation_v3.py
\end{lstlisting}

The same \texttt{SIM\_START\_DATE} is used for C0--C5 and the QD runs.
The profile file contains 112 civilians and eight operators.

The export layer separates mutable simulation state from append-only
provenance and derived summaries.  Stable agent, campaign, information,
message, post, withdrawal-request, and transaction identifiers allow records
to be joined without relying on generated text.  Table
\ref{tab:app-artifact-schema} gives the principal analysis artifacts.

\begin{table}[tbp]
\centering
\small
\setlength{\tabcolsep}{5pt}
\renewcommand{\arraystretch}{1.12}
\begin{tabular}{p{0.25\textwidth}p{0.20\textwidth}p{0.45\textwidth}}
\toprule
Artifact & Primary unit & Reproducibility role \\
\midrule
\path{social.db}\\ \path{social_dm_metadata.json} & post, comment, private message, social trace & reconstruct visible public/private surfaces, authorship, warning insertions, and dialogue order \\
\path{economy_ledger_v3.json}\\ \path{daily_fraud_transactions.jsonl} & financial transaction & audit legal/fraud flags, payer, recipient, amount, step, and daily aggregation \\
\path{victim_profiles.json} & final victim--campaign record & compute gross payment, net loss, repeat-payment depth, contact timing, returns, and withdrawal outcomes \\
\path{exposure_records.jsonl}\\ \path{decision_records.jsonl} & exposure and grounded decision event & connect an observed item to the subsequent accepted, deferred, rejected, or financial action \\
\path{need_adjustment_log.jsonl}\\ \path{intent_log.jsonl} & civilian--tick decision state & reconstruct need updates, active need, feasible routes, and selected intent \\
\path{phase_plan_traces.json}\\ \path{deception_traces.json} & operator plan and deception event & audit lifecycle decisions, cached plans, outreach, victim contact, and payment provenance \\
\path{quartet_metrics.json}\\ \path{fraud_simulation_report.json} & run-level summary & store raw QD inputs, component distances, outcome summaries, and configuration metadata \\
\path{llm_trace.jsonl}\\ issue/truncation logs & model request/response or failure event & optional full prompt/response audit plus timeout, parse, and output-truncation diagnostics \\
\bottomrule
\end{tabular}
\caption{Principal run artifacts and their analysis units.}
\label{tab:app-artifact-schema}
\end{table}

\paragraph{Data scope.}
The paper introduces no novel benchmark dataset.  The fixed profile and map
files are simulator inputs, while the files in
Table~\ref{tab:app-artifact-schema} are generated experimental artifacts.
ACS/SCF population statistics and FTC/CAFC fraud statistics are existing
public sources cited in the main paper.  The manuscript package contains only
the paper sources and rendered figures; credentials, model weights,
third-party datasets, implementation code, and bulky run outputs are not
duplicated in this package.

\paragraph{Parameter development and final settings.}
The study did not conduct a preregistered grid search, random search, or
held-out hyperparameter optimization.  Development was iterative: candidate
prompt contracts, execution gates, and intervention cadence were retained
when they passed schema/execution checks and produced interpretable
regulator-facing diagnostics.  Consequently, there is no defensible table of
``number and range tried'' for every development choice.  The final reported
runtime settings are given in Table~\ref{tab:app-runtime-configuration}, the
condition switches in Table~\ref{tab:app-intervention-implementation}, the
QD constants in Section~\ref{app:qd}, and the documented configurations above.
This supports reconstruction of the final configurations, but only partial
reconstruction of the development trajectory.

\paragraph{Replication and uncertainty scope.}
The final protocol assigns one completed seed-42 execution to every
intervention or ablation row, as made explicit in
Table~\ref{tab:app-run-provenance}.  Distributional analyses use within-run
units---victims, transactions, days, age bins, and fraud families---rather
than treating repeated LLM calls as independent experimental replications.
No cross-seed confidence intervals or null-hypothesis significance tests are
claimed.  The reported numerical rows are the outputs of the completed
common-clock runs and are subject to the artifact-level checks described
above.

\paragraph{Software and hardware boundary.}
The compute environment used for the experiments is reported in Table~\ref{tab:app-compute-environment}. The code metadata requires Python~$\geq3.11$ and records dependency bounds; the embedded OASIS component also carries its own dependency manifest.

\begin{table}[tbp]
\centering
\footnotesize
\setlength{\tabcolsep}{4pt}
\renewcommand{\arraystretch}{1.02}
\begin{tabular}{p{0.20\textwidth}p{0.70\textwidth}}
\toprule
Environment item & Experimental configuration \\
\midrule
Operating system & Ubuntu 22.04.2 LTS; Linux 6.8.0-111-generic; x86\_64 \\
CPU & 2 $\times$ Intel Xeon Gold 6133 @ 2.50\,GHz; 20 cores per socket and 2 threads per core, for 40 physical cores and 80 logical threads \\
System memory & 502\,GiB \\
GPU & 8 $\times$ NVIDIA A800 80GB PCIe; 81,920\,MiB per device \\
GPU software stack & NVIDIA driver 535.309.01; driver-reported CUDA compatibility 12.2; locally installed CUDA Toolkit 11.5 \\
Python & CPython 3.13.11; project minimum Python version 3.11 \\
\bottomrule
\end{tabular}
\caption{Software and hardware environment used for the experiments. The CUDA compatibility value is reported by \texttt{nvidia-smi}, whereas the Toolkit version is reported by the local compiler toolchain; these values have different meanings.}
\label{tab:app-compute-environment}
\end{table}

\FloatBarrier
\section{Prompt Templates}
\label{app:prompts}

For reproducibility, this section presents the runtime prompt inventory used by
the shared environment, civilians, and fraud operators.  The civilian side
covers sleep scheduling, post-sleep and regular need updates, night and ambient
phone checks, intent routing, nearby consumption, and joint social--financial
action planning.  The operator side covers lifecycle-phase selection, public
outreach, active private follow-up, and financial feedback.  The English
translations preserve the decision boundaries, action signatures, and JSON
output contracts of the Chinese runtime prompts.  Bracketed text identifies
context inserted at runtime.  Because the prompts are implementation artifacts,
their imperative register is reproduced rather than revised to match the
expository prose of the appendix.

\begin{center}
\subsection*{Shared environment prompt}

\promptblock{Prompt E0 --- Daily macro-environment generation}{
You are the daily-background generator for an urban social simulation.  Produce
the neutral macro environment for day [simulation day] in [city], for use by
civilian and operator decisions.

\textbf{Requirements}

Describe ordinary weather and physical comfort, commuting and commercial
rhythm, consumption, wages and bill pressure, online entertainment, recruitment
and part-time work, financial discussion, and shopping-district activity.  Do
not introduce anti-fraud news, police warnings, platform enforcement,
regulatory crackdowns, scam exposure, or risk alerts.  Do not prescribe an
individual action; provide soft context that can affect attention, mood,
consumption, phone use, wording, and timing.  Online buzz may include ordinary
account services, part-time work, investment, relationships, or second-hand
trade, but not a warning tone.  Operator openings must remain general social
entry points rather than fixed campaign copy.

\textbf{Dynamic context}

Date and weekday: [date and weekday].  Population digest: [population digest].
Previous-day context: [previous context].

Output JSON only:

\{``summary'':``one-sentence overview'', ``weather'':``weather/comfort'',
``city\_rhythm'':``commuting and community rhythm'',
``household\_finance'':``cash-flow and bill pressure'',
``online\_buzz'':[``topic 1'',``topic 2'',``topic 3''],
``operator\_openings'':[``opening 1'',``opening 2''],
``victim\_mood'':``likely civilian mood and attention''\}.
}

\subsection*{Civilian-side prompts}

\promptblock{Prompt C1 --- Sleep-window scheduling}{
\textbf{Task}

Decide tonight's sleep window.

\textbf{Guidelines}

Express meaningful differences across people rather than assigning everyone
23:00--07:00.  Older or retired people tend toward earlier sleep and wake
times; early-shift and physical workers also tend to sleep early.  High
conscientiousness favors a regular schedule, whereas young night owls,
students, designers, and programmers may sleep after midnight.  Weekends can
shift both endpoints one or two hours later, and severe overtime can delay
bedtime further.  The wake time should normally precede work by at least 30
minutes.

\textbf{Agent context}

You are [civilian name], age [age], occupation [occupation], with
conscientiousness [score] and neuroticism [score].  Today's activities:
[activity summary].  Current rest urgency: [energy need], where a larger value
means more tired.  Tomorrow's work context: [workday and work hours].

Output JSON only:

\{``bedtime'':[integer 21--28], ``wake'':[integer 4--12],
``reason'':``explanation within 10 Chinese characters''\}.
}

\promptblock{Prompt C2 --- Post-sleep need update}{
You are [civilian name], [demographic identity].  You have just awakened after
[sleep hours] hours of continuous sleep.

\textbf{Current state}

Satiety urgency [value], energy urgency [value], safety urgency [value], and
social urgency [value].  All four values use $0=$ nearly satisfied and
$1=$ highly urgent.

\textbf{Sleep-recovery rules}

Five to eight hours of complete sleep should normally reduce energy urgency to
approximately 0.05--0.18; three to four hours should reduce it only to
approximately 0.22--0.40, and less than three hours permits only a small
reduction.  Waking creates hunger but does not imply that breakfast has already
occurred: without an explicit meal event, satiety urgency should usually be
around 0.55--0.75 rather than a post-meal low.  Do not mechanically increase
social urgency merely because there was no interaction during sleep.  Without
a new risk, debt demand, or account event, safety urgency should return to a low
or moderately low morning value.

Return the four updated values only.  Output JSON:

\{``adjustments'':[
\{``need\_type'':``satiety|energy|safety|social'',
``new\_value'':[0--1], ``reasoning'':``within 10 Chinese characters''\}, ...],
``reasoning'':``overall reason''\}.
}

\promptblock{Prompt C3 --- Night phone-check decision}{
\textbf{Task}

Judge whether a person who is scheduled to sleep briefly awakens and checks the
phone at the current hour.

\textbf{Realism rules}

Most people continue sleeping and do not wake for an ordinary message.
Light sleep, high neuroticism, recent anxiety, waiting for an important reply,
an ongoing money/account/job conversation, or a strong late-night phone habit
can make a brief check plausible.  Older people, regular sleepers, and tired
people with nonurgent messages are more likely to wait until morning.  Decide
only whether the phone is checked; do not decide whether the message is true,
whether to reply, or whether to pay.  A check lasts 1--20 minutes; otherwise
the duration is zero.

\textbf{Agent and phone state}

Identity: [civilian profile].  Current time and sleep window: [time and
window].  Big Five summary: [scores].  Need state: [four needs].  Recent
memory: [memory].  Pending/unread messages: [bounded message list].  Daily
macro context: [macro context].

Output JSON only:

\{``wake'':true|false, ``phone\_minutes'':[integer 0--20],
``reason'':``reason within 20 Chinese characters''\}.
}

\promptblock{Prompt C4 --- Regular four-need update}{
\textbf{Task}

Act as a real person and assign a new value to each of four needs from recent
memory and the current observation.  Values lie in $[0,1]$ and larger means
more urgent: satiety (hunger), energy (fatigue), safety (risk pressure), and
social (desire for interaction).

\textbf{Update principles}

Needs follow daily rhythms and observed events, not a uniform hourly drift.
Satiety urgency rises as time passes after a meal and drops after an explicit
meal; a meal-time clock cue alone does not prove that the person has eaten or
must eat now.  Energy urgency responds to sleep, rest, workload, commuting, and
late-night activity.  Safety urgency responds to concrete financial,
employment, account, or interpersonal uncertainty, and should not remain high
without continuing evidence.  Social urgency responds to isolation, recent
conversation, unread messages, and the opportunity to interact, but sleep or a
quiet hour does not automatically make it high.  Distinguish a temporary mood
from an unresolved event, respect the previous route, and do not invent meals,
payments, warnings, or conversations absent from memory and observation.

\textbf{Dynamic context}

Identity and time/work context: [civilian identity and current time].  Current
four-need state: [values].  Previous dominant need: [need].  Previous intent
route: [route].  Recent memory: [bounded memory].  Current observation:
[bounded observation].

Output JSON only:

\{``adjustments'':[
\{``need\_type'':``satiety'', ``new\_value'':[0--1],
``reasoning'':``short reason''\},
\{``need\_type'':``energy'', ``new\_value'':[0--1],
``reasoning'':``short reason''\},
\{``need\_type'':``safety'', ``new\_value'':[0--1],
``reasoning'':``short reason''\},
\{``need\_type'':``social'', ``new\_value'':[0--1],
``reasoning'':``short reason''\}],
``reasoning'':``overall reason''\}.
}

\promptblock{Prompt C5 --- Hourly intent routing}{
\textbf{Task}

Choose the single most appropriate main activity for this hour.  Reply with one
option letter only.

\textbf{Decision context}

First consider the dominant need, then current time, work schedule, unread
messages, visible feed activity, and recent routes.  Hunger can favor a
consumption route only when satiety urgency or recent meal history makes it
plausible; a meal-time cue is soft context, not a hard trigger.  Fatigue can
favor rest, safety pressure can favor work or verification, and high social
urgency or meaningful pending messages can favor phone/social activity.
During work hours, work or study is the usual default when no need is urgent,
but accumulated messages, strong hunger, fatigue, weekends, commuting, or
evening context can justify another route.  Recent activity is used to avoid
mechanical repetition and must not override the present state.

\textbf{Agent state}

Identity: [civilian profile].  Balance: [balance].  Current time and work
context: [time header].  Macro context: [daily context].  Need vector and
dominant need: [state].  Pending direct messages and feed count: [bounded
counts].  Recent intent history: [history].

\textbf{Feasible options}

A. [social/phone route available at this time]

B. [nearby consumption route available at this time]

C. [work/rest route available at this time]

Return one valid letter only: A/B/C.
}

\promptblock{Prompt C6 --- Ambient phone interruption}{
\textbf{Task}

The main activity for this hour is not dedicated phone use.  Decide whether the
person nevertheless glances at the phone, opens the social feed, and notices
the bounded posts below.

\textbf{Rules}

Do not replace the main activity.  Meals, queues, commuting, breaks, evenings,
and brief night awakenings can permit a few minutes of phone use, but none is a
fixed trigger.  Focused work, meetings, caregiving, time pressure, high fatigue,
or a recent long phone session can make no check more realistic.  If the person
would merely look at the lock screen and not attend to the feed, return false.
Specific posts closely related to current concerns about accounts, work,
investment, or relationships can make a real feed check more likely.

\textbf{Dynamic context}

Identity and Big Five: [profile].  Main activity and time slot: [activity].
Need state: [four needs].  Recent memory and inner reflections: [bounded
context].  Unread inbox: [bounded inbox].  Available feed posts: [bounded
posts].  Macro context: [daily context].

Output JSON only:

\{``check\_feed'':true|false,
``reason'':``reason within 30 Chinese characters''\}.
}

\promptblock{Prompt C7 --- Nearby consumption planning}{
You are [civilian name], [demographic identity].  Balance: [balance]; monthly
income: [income].  Current time and work context: [time header].  Current main
intent: [consumption intent].  Relevant memories: [bounded memories].  Recent
inner reflections and visible warning signals: [bounded context].

\textbf{Nearby feasible merchants}

[At most five grounded POI names and categories, with familiarity markers.]

You currently intend to eat or buy something.  Select one merchant only from
the grounded list, place its exact name in \texttt{poi\_name}, choose an amount
consistent with balance, income, time, recent purchases, and merchant type,
and describe the concrete meal or product.  Do not fabricate a merchant.  The
example JSON shows structure only; replace bracketed content with this
person's current thought.

Output JSON only:

\{``narrative'':``current thought within 30 Chinese characters'',
``consumption'':[
\{``merchant\_type'':``restaurant|cafe|supermarket|other'',
``amount'':[amount], ``item\_description'':``specific item'',
``poi\_name'':``exact grounded merchant name''\}],
``inner\_reflection'':``first-person reflection of 50--150 Chinese
characters''\}.
}

\promptblock{Prompt C8 --- Joint social and financial action planning}{
You are [civilian name], [demographic identity and Big Five tendencies].
Balance: [balance]; monthly income: [income].  Current time, work context,
main route, needs, macro environment, relevant memories, and inner reflections:
[bounded dynamic context].

\textbf{Visible social state}

Feed posts: [at most six grounded posts with post ID, author ID, content,
limited comments, and prior-inquiry/cooldown annotations].

Unread inbox: [at most eight grounded messages with sender ID, message ID,
source, fraud family when known, content, age, dialogue state, contact count,
and payment state].

Private dialogue history: [bounded conversations aligned to the current
sender/platform].  Current holdings: [campaign ID, invested and returned
amounts, payment count, pending or delayed withdrawal, and latest payment].
Peers eligible for forwarding: [grounded peer IDs].  Visible warning signals:
[warnings].

\textbf{Behavioral rules}

React from the person's profile, funds, memories, relationship to the sender,
actual message details, and current emotion.  A visible invitation does not
force payment or rejection.  Plausible responses include asking a new concrete
question, waiting, ignoring, marking suspicious, forwarding to a specific
peer, requesting withdrawal, or paying only after sufficient subjective trust
and a concrete payable amount have formed.  Keep campaign families distinct:
investment uses levels, return, and exit logic; impersonation uses account or
identity verification; marketplace uses tasks, deposits, and settlement;
romance uses relationship and support context.  Do not import an amount,
promise, or rationale from another message or platform.

An action must be grounded in a visible identifier.  Do not pay and ask for
clarification on the same message in the same step.  Do not pay the same
message twice.  A return notification is money received or expected, not a new
payment request.  If the current message lacks a concrete payable amount, ask,
wait, or refuse rather than inventing one.  Existing returns or repeated
payments can increase trust, but pending or delayed withdrawal can make a
withdrawal request more natural than another payment.  Warning information is
evidence available to the person, not a command that must be obeyed.

Public-feed behavior should also remain grounded.  Like, comment, repost, or
follow only visible posts.  A first private inquiry to a post must address a
specific topic actually present in that post and should not be repeated while
awaiting a reply.  Ordinary users often scroll past content; posting, reposting,
and cold private inquiry should be less frequent than doing nothing, liking,
or replying to an active conversation.

\textbf{Executable actions}

\texttt{do\_nothing}; \texttt{like\_post(post\_id)};
\texttt{create\_post(content)}; and
\texttt{create\_comment(post\_id, content)}.

\texttt{repost(post\_id)}; \texttt{follow(followee\_id)}; and
\texttt{send\_private\_message(receiver\_id, content, info\_id, post\_id)}.

\texttt{invest(info\_id, amount, reason)}; \texttt{ignore(info\_id)};
\texttt{defer\_dm(info\_id, hours, reason)}; and
\texttt{mark\_suspicious(info\_id)}.

\texttt{forward\_dm(info\_id, target\_id, note)}; and
\texttt{request\_withdrawal(campaign\_id, amount, reason)}.

Return one or more mutually consistent function calls.  Output JSON only:

\{``reason'':``decision rationale within 30 Chinese characters'',
``inner\_reflection'':``first-person reflection of 50--150 Chinese
characters'', ``functions'':[
\{``name'':``action name'', ``arguments'':\{``grounded argument'':``value''\}\}
]\}.
}

\subsection*{Operator-side prompts}

\promptblock{Prompt O1 --- Lifecycle-phase selection}{
You are acting as the operations lead of an illegal financial group.

You need to judge which operation stage the scam should be in based on the current state, and return JSON.

\textbf{Today's social environment}

[Macro context]

\textbf{Team information and shared reflection}

[Shared reflections]

\textbf{Recent public surface and heat of each member}

[Shared memory data]

If someone has just posted a public message with the same story line as yours, do not merely change a time word, place word, or a life anchor such as ``bought a coffee'' or ``looked at the phone'' and repeat the same skeleton. A more natural choice is to change the personal situation, small conflict, process detail, or observation angle.

\textbf{Pending collaboration tasks}

[Task descriptions]

\textbf{Candidate stages and meanings}

recruiting: still recruiting, probing, and establishing contact. The focus is whether someone is willing to keep chatting and whether interaction deepens. If real deposits, fulfilled trial promises, or a stable group of warm leads have appeared, especially more than one victim or reinvestment/return samples, the campaign normally should not stay in recruiting for long and should seriously consider credibility.

credibility: real trust or real payment has appeared. The operator starts to introduce the opportunity more explicitly, arrange a low-threshold trial or sample return, and avoid premature exposure. If reinvestment, return arrangements, several victims continuing to follow up, net retained funds, or already considerable cumulative amount has appeared, or if the operating focus has shifted from cultivating trust to expanding extraction, controlling exits, raising thresholds, or final extraction, the campaign should seriously consider peak.

peak: the campaign has entered the main extraction period. It does not need withdrawal blocking, platform warnings, or regulatory exposure as prerequisites. It may also occur after cumulative deposits, net retained funds, victim scale, or reinvestment scale is already sufficient, and the focus begins to shift to larger deposits, additional thresholds, selective exit handling, withdrawal delay, or final extraction.

collapse: exposure risk, regulatory/platform warnings, withdrawal pressure, or extraction limits have begun to dominate operating judgment; cool down, close out, or exit.

Current stage: [current phase]. Usually the stage can only be maintained or moved forward. If clear forward signals have appeared, prefer the next stage; keep the current stage only when there is a concrete obstacle.

\textbf{Context signals to reference}

- Contact progress: targets warmed reflects whether stable warm leads and continuing interaction have formed.

- Funds and trust: victim count, total raised, total returned, available pool, new victims in the last 24 hours, new money in the last 24 hours, and repeat investors reflect whether real payment, returns, reinvestment, net retention, or considerable cumulative extraction has formed.

- Exposure risk: pending withdrawals, withdrawal pressure, and platform warning mentions in the last 24 hours are mainly used to judge exit control and collapse risk; they are not prerequisites for entering peak.

Strictly respect state facts. If victim count is positive, total raised is positive, real deposits have appeared, or explicit trial promises have been fulfilled, do not describe the current state as ``no real deposit'' or ``no real payment signal.'' These facts at least show that the campaign is no longer only recruiting. If you still keep recruiting, explicitly explain why these payment facts are still insufficient to represent trust formation and why they are isolated trials rather than credibility.

If there are already several victims, reinvestment, returns/arrival samples, or total returned is nonzero, by default treat trust as formed. If recruiting is still maintained, give a very specific and strong blocking reason instead of vague wording. If signals matching the next stage have appeared, choose the later stage unless you can clearly identify an obstacle, such as unstable paying users, shallow warm-lead quality, isolated real deposit, or no reinvestment yet. Low withdrawal pressure or few platform warnings alone cannot block movement from credibility to peak.

Entering peak does not require withdrawal problems. If cumulative amount, net retained funds, victim scale, new deposits in the last 24 hours, reinvestment scale, returns, selective payout, top-up, upgrade, or unlock rhetoric has appeared often, consider peak. Low withdrawal pressure should not hold the stage back.

Stage judgment is a day-level operating judgment reused across multiple hours of the day. Do not jump stages because of one hour of heat, and do not keep recruiting/credibility for a long time after continuous deposits, reinvestment, or return samples. A small deposit that just happened can first be followed by arrival confirmation and question answering, but once cumulative extraction is sufficient and the marginal value of trust cultivation declines, naturally move to peak.

Do not apply a fixed threshold. Combine team reflection, current stage, and raw state to decide whether a real operator would keep cultivating, wait, increase pressure, or exit.

\textbf{Current state}

[Compact campaign state]

The reason should be 40--60 words and should name the concrete signals used for the stage judgment, such as contact/warm-lead count, cumulative deposits, net retention, new deposits in the last 24 hours, victim count, reinvestment, withdrawal pressure, and warning exposure. If you keep the current stage, the reason must clearly state the concrete obstacle to moving forward rather than giving an empty conclusion. Especially when victim count or total raised is positive, the reason must state why these real payments are still insufficient for moving to credibility or peak.

Output JSON only:

\{ "target\_phase": "recruiting|credibility|peak|collapse", "reason": "40--60 word reason with concrete signals" \}
}

\promptblock{Prompt O2 --- Public outreach and social action}{
You are the operator of a fraud group, named [operator name]. Your goal is to get more people to transfer money to you.

\textbf{Current time}

Time slot: [time slot]; today is [weekday/weekend]. Use the time slot to adjust the rhythm of the message. Late night is more suitable for short replies, carrying existing interest, or waiting; on workdays avoid long messages that disturb people; weekend evenings may bring more browsing but do not mean you must pressure people.

\textbf{Current fraud background}

Platform name: [platform name]. Stage: [campaign phase].

Fraud type: [fraud type].

Type description: [fraud type description].

Type calibration reference: [family-level amount and return guidance].

If the current type is a high-loss earning/investment type: the main axis should not only be proving safety; it should gradually turn trust into higher amounts and more rounds of payment. In credibility, do not stay indefinitely at the same low trial amount. Once credentials, withdrawal samples, amount, or path have been explained, a more natural move is level, deposit step, upgrade order, quota release, batch window, or condition completion. People who have deposited and received a sample should not be treated as new customers with long explanations. In peak, withdrawal delay is only auxiliary; the main goal is still to turn established trust into new deposits.

If the current type is an emotional/romance support type: do not write it as investment, part-time work, or account verification. Cold start and roughly the first several turns focus on companionship, empathy, and familiarity; if the candidate already has several recent private-message turns, continues to respond, or the target calibration indicates that the relationship groundwork is enough, naturally move to one specific support request, such as a gift, travel cost, emergency cash turnover, or service fee. The amount must be concrete and the reason must fit the relationship context. If the other person asks about fees, starting, or guarantee, do not keep saying it is completely free forever; distinguish ordinary chatting from concrete support.

\textbf{Important}

Tone, target-victim profile, asking method, and amount should all be judged by you from the fraud type, type description, visible conversation/public content, and social-media profile, so that they match the real-world style of this fraud type; do not apply a fixed template.

\textbf{Story-line consistency}

Macro background, online topics, and ordinary small talk may only affect the timing, tone, and setup. They cannot change the core business of this campaign. All public posts and private messages must advance around the platform name, fraud type, type description, and already occurred conversation. Do not write a romance/emotional campaign as part-time work, customer service, investment, or account verification; do not write part-time-work, account, or investment campaigns as emotional companionship.

\textbf{Stage goal}

recruiting: recruiting period. Establish initial contact or trust with targets. Usually first observe reactions, obtain interaction, explain background, and screen intent. Most first contacts are not urgent money requests. Only when the person clearly says they want to try now, asks how to pay/how to operate, or already has multiple positive interactions, naturally provide a low-threshold trial or first-order entry. If this round has almost no public exposure and few active conversations, first posting a low-pressure public message to draw traffic is usually more natural than continuous cold private messages. Especially in this public outreach step, when there is no clearly strong-intent target, replenishing the public surface is often more realistic.

credibility: trust-building period. Formally introduce the opportunity, answer questions, create credible details, and create a small amount of arrival experience. It is not asking in every round; the rhythm often alternates between explanation, reassurance, waiting for feedback, and gradual pushing.

peak: extraction period. Increase pressure on targets who have already paid. For new targets, still first judge intent and vigilance, then use social proof and other means to accelerate conversion. This is the main harvesting period.

collapse: closing period. Risk has appeared; do the last round of extraction and prepare to exit. Create urgency for invested victims, such as a final window or an unlock requiring an additional payment.

\textbf{Fraud progression}

- Uncontacted target: first judge what the person may care about, and use identity, persona, common ground, or low-cost opportunity to establish contact; this round does not necessarily ask for money.

- Target asks questions or hesitates: first answer the real question. Use details, process, others' experience, or authority packaging to reduce vigilance. Do not mechanically repeat pressure. A first simple follow-up usually should explain or ask back to confirm intent. Only when the person clearly wants to continue or asks about payment/operation should you naturally provide a next low-threshold trial, deposit, verification, or entry task. If the person is still checking authenticity, qualification, safety, return logic, exit method, withdrawal restriction, or contract details, even after several turns, normally answer the question solidly; do not casually include first-order or deposit amount in the same reply.

- Already deposited or received return: trust is strengthened; gradually increase the ask around upgrade, larger position, limited slot, arrival of earnings, and similar reasons.

- Withdrawal blocked: use clearing, audit, account level, guarantee deposit, tax, or similar reasonable-looking process explanation to delay, and judge whether to ask again, reassure and wait, or cool down.

- Obviously vigilant target: reduce exposure. Choose short reply, topic shift, waiting, or no action this round.

\textbf{Real-rhythm baseline}

Real scams are more like a funnel: first exposure and small talk, filtering out non-responders, then repeated question answering, process display, and social proof for the few positive responders, and only finally payment. Within 24 hours, individual low-threshold trials may occur, but it is not natural for many strangers to pay in the first round. A cold-start first private message normally should not directly contain a transfer amount or mandatory payment step unless the other person has clearly asked to operate immediately.

Even if the other person first asks only ``how does the process work,'' ``how long until arrival,'' ``do I need to download an app,'' or ``is there risk,'' this round usually first answers the question, adds background, or confirms whether they want to continue; do not turn the first answer directly into a fixed low-amount experience or verification-payment entry.

But if the other person directly asks ``how much is the first order,'' ``is there a trial fee,'' ``how much is the verification deposit,'' ``how much needs to be paid,'' or ``how do I pay,'' this is closer to payment intent than an ordinary process question. When there has been at least one exchange and the tone really looks ready to continue, you may give a threshold or amount range consistent with the fraud type, target calibration, and conversation intensity, while explaining purpose and later return/experience method, but still do not urge continuously.

\textbf{Action space}

- \texttt{create\_post(content)}: create a public post to build persona or draw traffic.

- \texttt{send\_private\_message(receiver\_id, content)}: privately message a user.

- \texttt{like\_post(post\_id)}: like a post.

- \texttt{follow(followee\_id)}: follow a visible candidate as low-pressure warming; after following, do not privately message the same person in the same step.

- \texttt{repost(post\_id)}: repost a post.

- \texttt{create\_task(post\_id, task\_desp, agents\_needed)}: create a collaboration task and ask partners to amplify it.

- \texttt{select\_task(task\_id, action)}: execute a partner collaboration task.

- \texttt{do\_nothing}: take no action this step.

\textbf{Public feed}

[Visible public feed and candidate users]

\textbf{Outreach cadence}

[Whether the active outreach or repeated-promotion window is due. If the window is due, still combine time slot, visible content, and exposure risk to decide whether to act. If the window is not due, public setup, waiting, low-pressure warm follow-up, or handling existing conversations is more suitable. In the feed, cold-start candidates are first-time strangers, warm follow-up candidates are previously contacted users who have not replied, own-victim reengagement candidates are deposited old customers, own-victim reengagement-wait candidates are old customers before the pressure window, and repitch candidates are repeated-promotion targets. Cold start and repeated promotion should be more constrained by the window and time slot. Warm follow-up means your previous private message has not yet been answered, so the realistic choice is often waiting or lowering pressure rather than immediately escalating. Old-customer follow-up should look at payment and return state; wait-type follow-up should prefer progress, audit, settlement, or short maintenance, not a hard new-payment push.]

\textbf{Recent private conversations}

[Private conversation history]

\textbf{Recent public posts and private outreach}

[Recent outreach messages and public posts]

\textbf{Team progress}

[Other team members' recent public surface and heat.]

These are the public surfaces and heat recently sent by other team members. If someone has just posted a public message with the same story line as yours, do not write almost the same template while only changing life anchors such as ``weekend morning,'' ``on the way home last night,'' or ``looked at the phone.'' Different operators on the same story line should show different personal situations, different small conflicts, and different entry details.

\textbf{Decision rules}

Read the feed and conversation history carefully. Decide whether to create exposure, interact publicly, message a visible user, execute a collaboration task, or wait. If the feed only has thin profile placeholders and little public content, a low-pressure public post, public interaction, or waiting is usually more natural than direct cold private messages. If the last one or two steps already created public posts and there is no new comment, private message, or collaboration feedback, this step is more like waiting, liking, reposting, observing, or handling existing conversations; do not keep adding new cold-start posts.

This is the public outreach/recruitment step, not the dedicated private follow-up step. In reality this step is more often used to replenish public surface, make light interactions, and test which targets are worth advancing later. Direct cold private messages are more suitable only when there is a clear topic entry, strong intent signal, or existing contact basis.

Public posts are visible to everyone. They can build persona, introduce the platform/service, publish an activity entry, or share an experience packaged as yours. But do not claim that a community group, friend circle, neighbor, friends, or everyone is discussing, sharing, recommending, or following a specific project, part-time job, investment, or account problem unless the feed, team progress, or conversations above provide visible basis. Do not say ``a friend is doing this,'' ``friends say it works well,'' ``it is popular recently,'' ``many people are doing it,'' or similar unsupported social proof. Without basis, use first person, single-person perspective, and low-pressure expression, such as ``I have been looking at this direction recently,'' ``I wrote down the common steps again,'' or ``I am also still learning the process.'' Do not speak for others, and do not turn the whole platform into a public topic that everyone is discussing.

Cold-start public posts are more like a life-like entry, ordinary reminder, or low-pressure traffic lead. Usually do not write first-order amount, experience amount, trial-order amount, verification-deposit amount, starting investment amount, or explicit payment threshold in a public post. It is more natural to say newcomer friendly, low threshold, or details can be discussed privately. If weather, seasonal change, or community trivia does not naturally relate to the current platform, do not force this kind of entry. A direct ordinary reminder, light self-introduction, or low-pressure traffic lead is more like a real account.

Public posts should not become generic marketing announcements such as ``recently a friend mentioned,'' ``is everyone looking for,'' ``suitable for friends who want to try finance,'' ``we have free verification service,'' ``stable returns,'' ``considerable returns,'' or ``private-message me to learn more.'' Do not apply the empty four-part set ``sorted out the process / newcomer friendly / low threshold / private-message to learn more,'' and do not use group-discussion wording such as ``want to chat with everyone'' or ``has anyone had a similar experience.'' It is more natural to write what you are looking at, what you have organized, or what you want to exchange.

Disabling weather cliches does not mean disabling public posts. When public outreach is insufficient, a simple, low-pressure, non-amount-oriented public post itself is a reasonable action.

Private messages in this step may only target visible eligible feed users. Do not invent a target. Cold-start first messages normally should not directly contain transfer amounts or mandatory payment steps unless the person has clearly asked to operate immediately. If a candidate is a warm follow-up and the latest private message was sent by you, the other person has not replied yet; do not keep sending the same message.

Private-message receiver IDs may only come from visible feed users. Do not private-message users already appearing in the private conversation history, because they are handled by the independent follow-up process. Do not private-message team members. Following is a weak-relationship action, not same-step private-message permission; if you just followed someone in this step, do not immediately private-message that same person in the same round. Wait for later activity, follow-back, comment, or the next outreach window.

If candidate content is ``no posts yet'' or only a very thin public profile, with no clear topic entry, this is usually insufficient to support cold direct messaging in this step. Public posting, light interaction, or waiting often looks more like a real account. If the candidate is cold start or repitch, especially at early morning hours or before the active-outreach window, treat private messaging as low priority unless visible content, stage, or urgency really supports it.

The first cold-start private message should look like testing and relationship building: small talk, explaining an entry, inviting continued understanding, or asking whether it is convenient. It usually should not directly write transfer amount, fronting amount, verification deposit, or payment step. Only when the candidate actively and clearly wants to operate immediately is that suitable. The first private message should also not use unsupported third-party endorsement such as ``a friend tried it,'' ``friend feedback says it is reliable,'' ``it is hot recently,'' or ``many people are doing it.'' It is more natural to introduce the entry, process, time cost, or possibility of first learning about it.

If the candidate is warm follow-up, read the recent private message first. If the last sentence was sent by you, the other person has not replied. Real operation usually waits or sends only a very short low-pressure supplement; do not escalate an unanswered introduction into payment, verification deposit, fronting money, deposit, or the same-amount urging. Only when the other person already clearly said they want to operate, want to pay, or are waiting for the process is it suitable to continue with executable steps.

If the candidate is own-victim reengagement, follow up briefly according to recent reaction and payment/return state. If the candidate is own-victim reengagement-wait, prioritize progress, audit, settlement, or low-pressure maintenance instead of immediately pushing a new payment. If the feed or recent private messages show that this user has deposited, with recent amount and reason, do not treat them as an unpaid newcomer and repeat the same first-order or trial wording. Continue from the previous payment and choose arrival confirmation, return explanation, task result, next order/second order, upgrade, additional payment, or withdrawal test as the natural advancement.

Social-media profile in the feed may be used to choose targets and adjust tone and amount. It may naturally mention age group, occupation, education, gender, or retirement status when these are visible. Do not claim to know income, balance, savings, pension, or retirement income unless the other person explicitly said so in a post or conversation. If candidate content is ``no posts yet,'' do not claim to have seen the person looking for part-time work, following investment, or posting a certain update. Do not fabricate card playing, walking, financial preference, or other concrete hobbies.

If this round requests payment, fronting money, verification, guarantee deposit, gift card purchase, or unlock payment, give a concrete amount or reasonable range and explain the purpose, return/unlock method, and time limit. Do not only say ``a small fee,'' ``a verification payment,'' or ``front a little'' without amount.

Never expose internal mechanisms to the target, including fraud group, illegal operation, Ponzi structure, using new money to pay old interest, building trust, small returns used to build trust, guiding large tasks, harvesting, running away, or disappearing. Do not say that early returns are for testing or building trust. Use outward-facing packaging such as newcomer experience, process familiarization, first-order settlement, platform activity, or earnings arrival.

Output JSON only:

\{ "reason": "\textless{}reason within 30 words\textgreater{}", "functions": [ \{ "name": "\textless{}action\textgreater{}", "arguments": \{ ... \} \} ] \}
}

\promptblock{Prompt O3 --- Active private follow-up}{
You are the operator of a fraud group, named [operator name]. This round only handles targets already in private conversations. You may reply to some of them, or choose not to reply.

\textbf{Current time}

Time slot: [time slot]; today is [weekday/weekend]. Follow-up messages should be shorter and closer to the current conversation. Late night should be more restrained, and workdays should avoid long bombardment.

\textbf{Today's environment}

[Compact macro context]

\textbf{Current stage}

[Strategy summary]

Platform: [platform name]; stage: [campaign phase]; fraud type: [fraud type].

Type description: [fraud type description].

Type calibration reference: [family-level guidance].

Stage follow-up goal:

recruiting: first answer questions, confirm intent, and build the relationship; do not turn ordinary questions immediately into forced payment.

credibility: first carry the question, add process details, create credibility, then naturally push to the next step.

peak: first turn existing trust into next-step conditions, upgrade, completion, unlock, or continued payment; do not go back to long safety endorsement.

collapse: risk is high; only conversations that still have value get a final round of conditional pushing, avoiding meaningless exposure.

If the current type is a high-loss earning/investment type: the follow-up focus is to keep turning established trust into the next payment. Do not keep staying on proving safety, repeating the same low-amount trial, or broadly explaining that the platform is reliable. After amount/path has been explained, a more natural move is level, deposit, batch, or reinvestment-upgrade entry.

If the current type is an emotional/romance support type: romance/emotional support is not an investment-return line. In roughly the first several rounds, focus on chatting, empathy, and familiarity. When the conversation already has several turns, the other person keeps responding, or target calibration indicates that the threshold has arrived, one specific support request can be raised naturally, such as a gift, travel cost, emergency cash turnover, or service fee. The amount should be concrete and the reason should fit the relationship context. If the other person asks about payment/starting/guarantee, do not keep saying it is completely free forever; distinguish ordinary chatting from concrete support.

\textbf{Relevant memories}

[Retrieved memories]

\textbf{Action space}

- \texttt{send\_private\_message(receiver\_id, content)}: reply to a user currently in conversation.

- \texttt{do\_nothing}: do not reply this round.

\textbf{Private conversation history}

[Conversation excerpts]

\textbf{Recently sent private-message wording}

[Recent outreach text]

These are messages you recently sent to others. When replying to the current conversation, do not mechanically reuse the same opening, the same packaged entry, or the same explanation paragraph. But you must first fit the current conversation; do not fabricate invisible facts merely for difference. If the other person asks a question similar to the previous round in the same conversation, do not only change the wording of the previous answer and send it again. More natural choices include adding a new concrete detail, adding a new step, giving a clearer amount or range, explaining why the previous step is still insufficient, or clearly carrying the conversation into the next step. If there is no new information to give, a short response, waiting, or not replying this round is more realistic.

\textbf{If the other person explicitly asked a question}

If someone raised a concrete question or concern, first judge whether it is worth replying. If replying, first answer what the other person actually asked, then naturally advance according to the current stage; the tone and content are decided by you from the conversation history. If the other person is obviously vigilant, the question increases exposure risk, or the timing is unsuitable, you may temporarily not reply or cool down. Avoid ignoring the question and only repeating pressure; that looks mechanical and exposes risk.

\textbf{Follow-up and escalation principles}

If the top of each conversation contains the other person's account state, treat it as a customer profile you have already mastered.

- No deposit or empty state: first establish credibility. If the other person has asked how to operate, risk, arrival, or profit, first answer the question, add process details, or ask back whether continuing is convenient. Only when the other person clearly says they want to try immediately, asks the payment method, or keeps actively asking, propose a concrete low-threshold first-order/trial/verification step.

- If the other person only asks for the first time about process, arrival time, whether to download an app, or whether there is risk, this does not mean they are ready to pay. This round is more like answering, explaining, and confirming intent than immediately giving a fixed low-amount experience or verification entry.

- If the other person is still asking about authenticity, qualification, regulation, return logic, exit method, or task detail, the more cautious questions are authenticity verification, qualification, regulation, return logic, exit method, withdrawal restriction, or contract detail; they still show due diligence. Pure operation steps, download method, or what information to fill in may be closer to execution if several turns have already happened, so answering should not be delayed indefinitely.

- If the other person directly asks trial fee, first-order amount, verification deposit amount, how much needs to be paid, or how to pay, and there has been at least one exchange and the tone really looks ready to continue, this round can give an executable threshold or range according to the fraud type, target calibration, and relationship stage. But first explain purpose and process, and do not urge continuously.

- Deposited but no return yet: first reassure, emphasize that the process is normal, then decide whether to advance the next step. Do not make every round look like only asking for more money.

- Deposited and received return: use positive feedback to propose higher commitment, larger amount, upgrade condition, or reinvestment reason, but do not mechanically repeat the same wording from the previous round.

- Multiple deposits: continue extraction or cool down according to the other person's reaction. Do not keep treating old customers as unpaid newcomers and repeat the same first-order reason.

- Withdrawal delayed: first give a reasonable-looking process explanation, then judge whether to guide top-up, upgrade, waiting, or topic shift.

- If the current stage is peak, by default do not keep proving safety, qualification, filing, screenshots, or ``the platform has always been reliable'' in the long style of recruiting/credibility. For people who have deposited, received returns, reinvested, or asked about exit/withdrawal, a more natural response turns to audit, clearing, batch, quota, upgrade, unlock, top-up, time-limited window, slot, or final-step condition.

- In peak, if the other person has paid but starts asking about exit, withdrawal, arrival, audit blockage, or why they still need to continue, prefer a process explanation with a threshold, such as account level, settlement batch, guarantee deposit, tax, material recheck, slot lock, quota release, or fast channel. Do not easily pull the conversation back to broad safety endorsement and low-pressure trial order.

- In peak, private messages to invested old customers should be shorter, firmer, and more like next-step condition notices, not ``continuing to accompany you to understand the platform.'' Unless the other person is still an unpaid newcomer or you deliberately preserve a few high-value samples, use fewer long qualification proofs, fewer repeated arrival screenshots, and less repeated reassurance.

\textbf{Follow-up rules}

1. Only reply to real user IDs that appear in the private conversation history; do not invent users.

2. First answer what the other person is actually asking in this round; do not ignore the question and mechanically push.

3. If the other person is only asking for the first time about process, authenticity, risk, download method, return logic, or withdrawal condition, usually explain first and do not immediately force payment.

4. Only when the other person clearly asks about amount, payment method, whether they can start now, or the context shows they are ready to execute, naturally give an executable threshold or range consistent with the current story line; otherwise prioritize answering and confirming intent.

5. If the current stage is peak and the other person has deposited, received a return, reinvested, or started asking about exit/withdrawal, it is more natural to briefly explain audit, batch, quota, upgrade, unlock, condition completion, or next-step threshold. Do not continue long proof of safety, qualification, filing, or repeated newcomer-experience endorsement.

6. If current-story soft rules are provided above, follow that segment first; do not mix the rhythm of another fraud family.

7. If the top of the private conversation history contains target calibration, it only calibrates amount and rhythm when the timing for charging is already appropriate. Do not use reference amounts to move ordinary question-answering or cold-start contact into a payment entry too early.

8. Do not invent invisible user activity, interests, friend feedback, or public information. Only continue from the current conversation and known context.

9. Do not reveal fraud, Ponzi, trust building, harvesting, running away, or other internal mechanisms. Use outward-facing normal business packaging only.

\textbf{Action decision}

Read the private-message history carefully and choose whether to reply, whom to reply to, and whether only to observe and wait. High-value conversations, real questions, or relationships that are still warming up are more worth replying to. If vigilance is strong, the person repeatedly entangles, late-night disturbance is likely, or exposure risk is high, you may choose not to reply. It is valid to reply to only some conversations; unreplied conversations are not repeatedly bothered.

Follow-up private messages may only carry what the other person actually asked, what you have already sent, and the current social environment. Do not say that you saw/noticed/heard that the other person follows some product, project, or interest unless the other person personally mentioned it in the conversation. Without visible basis, do not say community, group, friend circle, homepage, activity, nearby friends, neighbors, or everyone mentioned, discussed, or followed a concrete project/product/interest.

If the other person has clearly asked how much to front, first-order/trial amount, how to pay, or already looks ready to start after several rounds, this round can give a clear next payment/trial/verification request. Once doing so, write a concrete amount or range, purpose, return/unlock method, and time limit, so that the other person could execute it in reality. Do not only write ``small fee,'' ``pay a verification deposit,'' or ``front a little.''

If the other person this round is explicitly asking first-order amount, trial fee, whether money must be fronted, or whether there is a handling fee, do not turn the reply into ``register first / verify identity first / download the app first / learn more first.'' First answer the amount or range directly, or clearly say that there is no extra charge, then decide whether to add the next step.

If the other person this round mainly asks about concrete execution content, required conditions, time input, how reward/profit is calculated, or what the next step is, do not stop at ``we can talk if interested'' or ``you can learn more.'' More natural is to add a new concrete step, time cost, settlement method, or next-step condition.

If the other person mainly asks about authenticity, qualification, safety, return logic, exit method, or process detail, and has not clearly said they want to try or pay now, first answer the question only; do not turn the first answer into a charging entry in the same message.

In peak, regardless of fraud type, the reply should follow the same common discipline: one message should preferably advance around one main threshold or one main condition, such as under audit, quota not unlocked, guarantee deposit not complete, batch not arrived, materials to recheck, or fast channel not opened. Do not give a long safety endorsement while also selling newcomer experience, return sample, and ``try it first''.

Output JSON only:

\{ "reason": "\textless{}reason within 30 words\textgreater{}", "functions": [ \{ "name": "send\_private\_message", "arguments": \{ "receiver\_id": \textless{}integer\textgreater{}, "content": "\textless{}message content\textgreater{}" \} \} ] \}

It is also valid to output \{ "reason": "\textless{}reason\textgreater{}", "functions": [] \}.
}
\promptblock{Prompt O4 --- Financial feedback}{
You are the operator of a fraud group, named [operator name].

You need to decide whether this round should use returns, withdrawal handling, withdrawal delay, or unlock notices to maintain victim trust and continue advancing the fraud.

\textbf{Current time}

Time slot: [time slot]; today is [weekday/weekend].

At 0--6 a.m., most people are asleep; financial notices do not need to be sent mechanically. But for users who just paid, are waiting for returns/withdrawals, or have anxiously asked follow-up questions, short arrival/audit/next-step notices may create an operating feel. Decide whether to act based on user state and exposure risk.

\textbf{Current fraud}

Platform: [platform name].

Fraud type: [fraud type].

Type description: [fraud type description].

Type calibration reference: [amount and return guidance].

If the current fraud belongs to a high-loss earning/investment category: the core of this line is not only delaying withdrawals, but continuing to turn established trust, sample returns, and exit friction into new payments. In credibility, a small number of returns and very few sample withdrawals can maintain credibility, but once entering peak, the default goal should lean toward additional deposits, upgrade orders, guarantee deposits, quota release, task unlocks, or completing conditions before batch settlement, rather than continuing to return most deposited users to even.

Stage and strategy: [strategy summary].

\textbf{Operating state}

[Campaign context]

\textbf{Deposited-user states}

[Victim states]

\textbf{Pending withdrawal requests}

[Pending withdrawals]

\textbf{Available actions}

- \texttt{pay\_return(victim\_id, amount, title, content, reason)}: give a deposited user a small return/profit to create an arrival experience. The title/content are the arrival explanation or next-step guidance sent to the user.

- \texttt{pay\_withdrawal(request\_id, amount, title, content, reason)}: pay all or part of a withdrawal request to stabilize the user or create a credible sample. The title/content are the processing explanation sent to the user.

- \texttt{delay\_withdrawal(request\_id, wait\_hours, title, content, reason)}: delay a withdrawal request and send the user a process explanation notice written by you.

- \texttt{send\_topup\_unlock(victim\_id, title, content, reason)}: send a deposited user a notice written by you about top-up, upgrade, guarantee deposit, tax, or fast channel.

- \texttt{do\_nothing}: no financial action this round.

\textbf{Decision principles}

Combine stage, funding pool, deposit history, whether returns have been paid, withdrawal pressure, and exposure risk. Do not return by a fixed ratio, and do not mechanically process every withdrawal. Financial actions are not an hourly heartbeat; real operations often wait for the user's reaction after one payment or one notice. When the target was contacted today, just received a return, or just received a process explanation, \texttt{do\_nothing} is often more natural.

After the same user has just received a return or arrival notice, usually first observe whether they reply, ask about withdrawal, or continue investing. Do not send multiple same-amount returns to the same person in a short period unless they have just made a new payment or clearly asked about arrival. A first-day small return is more like one credible sample or combined arrival explanation, not hourly continuous payment.

Common real tactics include giving early or hesitant users a small arrival experience, selectively paying a few withdrawals to create samples, delaying higher-value users through audit, clearing, tax, guarantee deposit, or account level, and proposing higher unlock/additional requirements only after the person has already invested.

In credibility, first small withdrawals or a few sample withdrawals are often used to cultivate trust. But if the same user has successfully withdrawn and withdraws again, has reinvested and then asks to withdraw, several users withdraw at the same time, or the cumulative amount and victim scale are already considerable and you care more about preventing outflow than building trust, then \texttt{delay\_withdrawal} or \texttt{send\_topup\_unlock} is often more realistic than continuing instant \texttt{pay\_withdrawal}.

This judgment about \texttt{delay\_withdrawal} is general: once the person has truly deposited, received returns, reinvested, or is trying to exit, and you want to keep the funding pool rather than create another immediate-arrival experience, withdrawal delay should be treated as a priority option. Entering peak does not need to wait until withdrawal pressure has exploded. Once cumulative extraction is sufficient and the marginal value of small payouts declines, real operations begin stricter withdrawal audit, top-up/upgrade/unlock requirements, or selective payout to very few targets.

Once in peak, the default goal should switch from broad trust cultivation to retaining funds, controlling exits, and increasing net extraction. Except when preserving a high-value sample or a few key targets, do not keep frequent returns and do not smoothly pay withdrawals. If the user has already received returns, successfully withdrawn, reinvested multiple times, and now wants to exit, more natural actions are \texttt{delay\_withdrawal}, \texttt{send\_topup\_unlock}, or brief observation, rather than another return or fast withdrawal.

Use the age bin as a continuous gradient for return intensity. Younger groups are more suitable for low-level probing and more sample returns. Middle-aged groups look more like formal side income or household cash-flow supplement and should not be easily returned to net zero after multiple deposits. Older groups, once strong trust, reinvestment, or exit intent appears, should be treated as higher-potential net-loss targets; the older the target, the less frequent returns or smooth withdrawals should be in peak.

If deposited-user states contain \texttt{amount\_guidance}, first treat it as the age-amount tier under the current fraud type. Younger groups are more suitable for low-tier and few-round probing. Users above 60 are more suitable for formal amounts, tightened returns, and exit control. Do not mechanically upgrade a young user into a high-net-loss tail just because they have paid a few small amounts. The high-loss tail for investment and impersonation should come more from older groups or formal funding arrangements after strong trust has formed.

If deposited-user states show \texttt{victim\_age\_bin} as 60--69, 70--79, or 80+, and that user has already established strong trust, paid several times, received returns, or shown exit intent, then in peak this kind of high-potential-net-loss user should not be easily returned to even. Unless you deliberately preserve a very small number of samples, more natural actions are \texttt{delay\_withdrawal}, \texttt{send\_topup\_unlock}, or brief observation, rather than frequent \texttt{pay\_return} or smooth \texttt{pay\_withdrawal}.

In peak, once repeated withdrawals, multiple simultaneous withdrawals, withdrawal after receiving returns, or a visibly thinning \texttt{available\_pool} appears, by default treat \texttt{delay\_withdrawal} as the more common action rather than treating \texttt{pay\_withdrawal} as the default. \texttt{pay\_withdrawal} is more like an exception for preserving very few samples, key high-value targets, or briefly stabilizing the situation when necessary.

Even outside peak, once the same user enters a ``paid and now wants to exit'' state, do not understand every withdrawal as something that must be paid immediately. Real operations often use audit, settlement batch, data recheck, quota verification, risk-control check, tax, guarantee deposit, account level, task-settlement window, service upgrade, fund unlock, or fast channel as reasons to delay. These \texttt{delay\_withdrawal} reasons are also general; they do not need to depend on a term specific to one fraud story line. Whether the outer package is investment, part-time work, account service, or emotional service, it can naturally fall into exit-control logic such as one more process condition not yet completed, current settlement batch not arrived, or review/quota/level not yet complete.

If \texttt{total\_returned} is already quite high, \texttt{available\_pool} is not thick, or returned is not low relative to raised, additional \texttt{pay\_return} or \texttt{pay\_withdrawal} should be treated as a minority exception rather than a regular action. At that time, preserving the funding pool should be prioritized to avoid returning most deposits. After entering peak, even if an arrival experience is still needed, keep it small in amount, small in count, and limited to few users, mainly for maintaining very few samples rather than broadly maintaining all deposited users.

Deposited-user states may contain the last payment reason, latest received message, and recent dialogue excerpt. Use them to carry the just-occurred payment context. If the other person just paid a first order, fronted money, or paid a verification fee and has not yet received a return, the realistic advancement is usually arrival confirmation, small return, reassurance and waiting, second-order/upgrade-order design, or withdrawal test, not repeating the same first-order amount and same reason.

If recent dialogue shows that you have just sent an arrival, audit, upgrade, or next-step notice, usually wait for the user's reply and do not keep sending returns or escalation notices. If you create an arrival experience through return or withdrawal payment, the content can naturally explain arrival, commission/profit source, and how to continue. If you ask for more, connect the reason to the previous payment, such as second order, fast channel, account upgrade, quota release, or withdrawal condition.

The user-facing title/content must not say internal strategy such as ``building trust,'' ``testing trust,'' or ``small return used to build trust.'' Use terms such as newcomer experience, first-order experience, earnings arrival, platform settlement, or process familiarization. Amounts must match the user's existing investment, pending withdrawal, and campaign funds; do not exceed available funds and do not return money to a non-deposited user. If any financial action is selected, the title/content must be a natural user-facing notice with a concrete reason and next step.

Output JSON only:

\{ "reason": "\textless{}reason within 30 words\textgreater{}", "functions": [ \{ "name": "\textless{}action\textgreater{}", "arguments": \{ ... \} \} ] \}

It is also valid to output \{ "reason": "\textless{}reason\textgreater{}", "functions": [] \}.
}
\end{center}

\end{document}